%% file: main.tex
\documentclass[]{fairmeta}

\title{Beyond Reasoning Gains: Mitigating General-Capability Forgetting  in Large Reasoning Models}

\author[1,2,*]{Hoang Phan}
\author[1]{Xianjun Yang}
\author[1]{Yuanshun Yao}
\author[1,3,*]{Jingyu Zhang}
\author[1]{Shengjie Bi}
\author[1]{Xiaocheng Tang}
\author[1]{Madian Khabsa}
\author[1]{Lijuan Liu}
\author[1]{Deren Lei}
\affiliation[1]{Meta Superintelligence Labs}
\affiliation[2]{New York University}
\affiliation[3]{Johns Hopkins University}
\contribution[*]{Work done at Meta}
\input{math_commands.tex}

\input{macro}

\input{sec/0.abstract}

\date{\today}

\usepackage{microtype}

\usepackage{inconsolata}
\usepackage{hyperref}
\usepackage{url}
\usepackage{multirow}
\usepackage{epsfig}
\usepackage{graphicx}
\usepackage{amsmath}
\usepackage{amssymb}
\usepackage{graphicx}
\usepackage{amssymb}
\usepackage{booktabs}
\usepackage{amsthm}
\usepackage{amsfonts}
\usepackage{algorithm}
\usepackage{algorithmic}

\usepackage{nicefrac}       
\usepackage{microtype} 
\usepackage{multirow}
\usepackage{wrapfig}
\usepackage{caption}
\usepackage{subcaption}
\usepackage{wrapfig} 
\makeatother
\usepackage{xcolor}
\usepackage{color, colortbl}

\correspondence{Hoang Phan  \email{hvp2011@nyu.edu}, Deren Lei \email{deren@meta.com}.}
\metadata[Code]{\url{https://github.com/VietHoang1512/recap}}

\newcommand{\PyComment}[1]{\ttfamily\textcolor{commentcolor}{\# #1}}  
\newcommand{\PyCode}[1]{\ttfamily\textcolor{black}{#1}} \newtheorem{theorem}{Theorem}
\newtheorem{lemma}{Lemma}
\newtheorem{proposition}{Proposition}

\definecolor{forestgreen}{RGB}{0,139,0}
\definecolor{violetpurple}{RGB}{126,63,242}
\definecolor{magentaaccent}{RGB}{221,0,116}
\definecolor{indigoblue}{RGB}{63,81,181}
\definecolor{commentcolor}{RGB}{110,154,155} 
\definecolor{carolinablue}{rgb}{0.6, 0.73, 0.89}
\begin{document}

\maketitle

\input{sec/1.introduction}

\input{sec/2.relatedwork}
\input{sec/3.background}
\input{sec/4.method}
\input{sec/5.experiment}
\input{sec/6.conclusion}

\bibliographystyle{assets/plainnat}
\bibliography{main}

\newpage
\beginappendix

\input{sec/appendix}

\end{document}

%% file: math_commands.tex
\usepackage{amsmath,amsfonts,bm}

\def\eqref#1{equation~\ref{#1}}

\def\1{\bm{1}}

\DeclareMathAlphabet{\mathsfit}{\encodingdefault}{\sfdefault}{m}{sl}
\SetMathAlphabet{\mathsfit}{bold}{\encodingdefault}{\sfdefault}{bx}{n}



%% file: macro.tex
\definecolor{lightgreen}{RGB}{201,242,155}
\definecolor{darkgreen}{RGB}{0,170,136}
\definecolor{na}{gray}{0.9}

%% file: sec/0.abstract.tex
\abstract{

\label{sec:abstract}
Reinforcement learning with verifiable rewards (RLVR) has delivered impressive gains in mathematical and multimodal reasoning and has become a standard post-training paradigm for contemporary language and vision-language models. However, the RLVR recipe introduces a significant risk of capability regression, in which models forget foundational skills after prolonged training without employing regularization strategies. We empirically confirm this concern, observing that open-source reasoning models suffer performance degradation on core capabilities such as perception and faithfulness. While imposing regularization terms like KL divergence can help prevent deviation from the base model, these terms are computed on the current task and therefore do not guarantee preservation of broader knowledge. Meanwhile, commonly used experience replay across heterogeneous domains makes it nontrivial to decide how much training emphasis each objective should receive. To address this, we propose RECAP—a replay strategy with dynamic objective reweighting for general knowledge preservation. Our reweighting mechanism adapts online using short-horizon signals of convergence and instability, shifting the post-training focus away from saturated objectives and toward underperforming or volatile ones. Our method is end-to-end and readily applicable to existing RLVR pipelines without training additional models or heavy tuning. Extensive experiments on benchmarks using Qwen2.5-VL-3B and Qwen2.5-VL-7B demonstrate the effectiveness of our method, which not only preserves general capabilities but also improves reasoning by enabling more flexible trade-offs among in-task rewards.
}

%% file: sec/1.introduction.tex
\section{Introduction}
\label{sec:introduction}



Large Language Models (LLMs) and Vision-Language Models (VLMs) have demonstrated remarkable general-purpose capabilities \citep{achiam2023gpt4, yang2023dawn}, yet strengthening their proficiency in complex reasoning remains a key frontier of research. Reinforcement Learning with Verifiable Rewards (RLVR) \citep{shao2024deepseekmath}, an extension of Reinforcement Learning from Human Feedback (RLHF) \citep{ziegler2019fine,ouyang2022instructgpt}, has emerged as a powerful paradigm for this purpose. By providing explicit reward signals such as exact-match correctness, format adherence, and answer brevity, RLVR has been applied to instruction following, STEM problem solving, code generation, and logical reasoning \citep{lightman2023let, peng2025verif}, resulting in large performance gains on benchmark scores, leading to headlines that language models can “learn to reason” \citep{guo2025deepseekr1}.


Despite strong headline gains, RLVR exhibits recurring failure modes, prompting questions about whether current pipelines genuinely expand reasoning abilities \citep{shojaee2025illusion}. For example, exploration and diversity collapse occur when on-policy finetuning overly narrows the policy distribution—raising Pass@1 while reducing Pass@k and solution-path diversity \citep{yue2025does, dang2025assessing}. Likewise, outcome-only rewards introduce sparse credit assignment and instability, and not every task can be naturally cast as a reinforcement-learning problem (e.g., translation, summarization, or captioning). 
In addition, strict answer formats and format-sensitive graders may conflate genuine reasoning improvements with mere format compliance, even introducing evaluation artifacts \citep{petrov2025proof}. Recent studies report that many RL-trained models underperform even their base models in standardized evaluations, where formatting-reward-only baselines can degrade original performance even more severely \citep{prabhudesai2025maximizing}. This suggests that the format reward may be underoptimized, yet optimizing it can also cause forgetting of mathematical capabilities \citep{llmrl2025incorrect}.

Another critical yet underexplored issue in RLVR is that optimizing for a narrow set of targeted  rewards can lead to regression in general capabilities acquired during pretraining. Although models become proficient in following formatting requirements and solving reasoning tasks, they simultaneously exhibit increased hallucinations \citep{jaech2024openai,yao2025reasoning} and become more vulnerable to jailbreak attacks \citep{lou2025think, yao2025mousetrap}. These results suggest that reasoning-oriented post-training can improve reasoning while trading off non-target competencies (e.g., perception, safety, factual grounding), especially when RL training is prolonged without explicit regularization \citep{liu2025prorl}.

\begin{figure}[t]
    \centering
     \includegraphics[width=.97\textwidth]{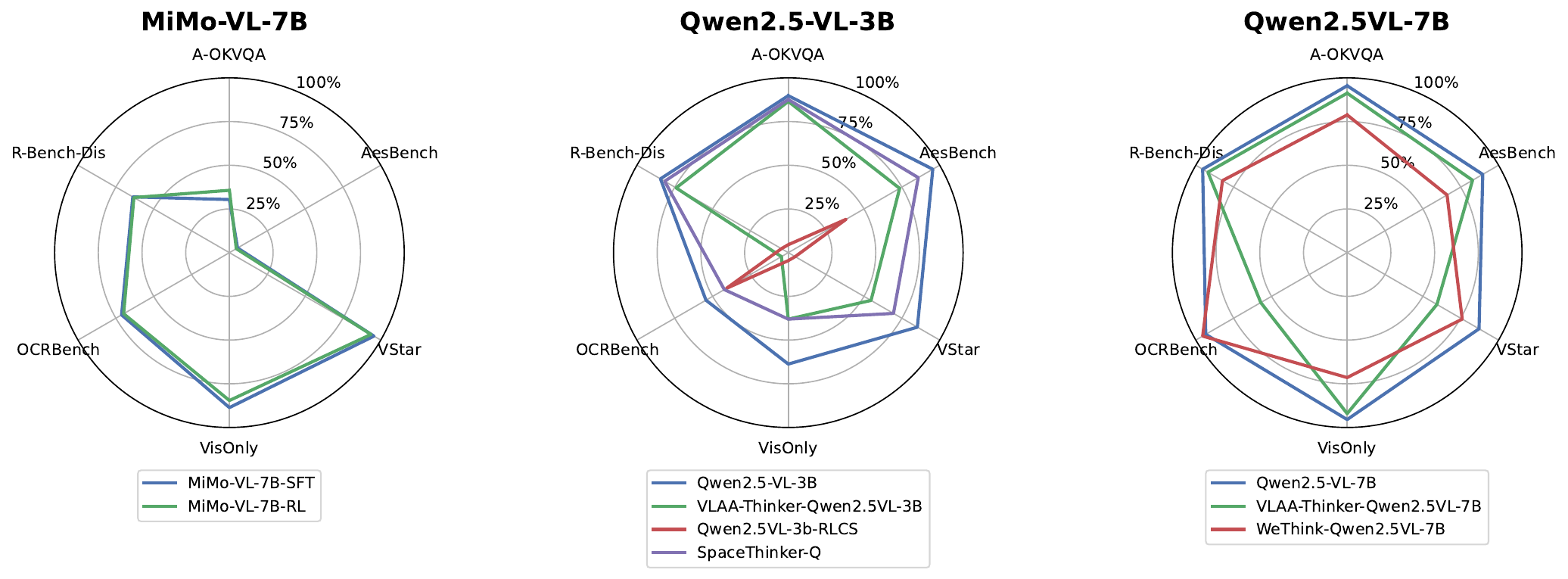}
    \vspace*{-2mm}
    \caption{
    \textbf{General-capability comparison of base VLMs (\textcolor{indigoblue}{blue}) and their reasoning-tuned variants (\textcolor{forestgreen}{green}/\textcolor{violetpurple}{purple}/\textcolor{magentaaccent}{red}) on six representative non-reasoning benchmarks (higher is better).}
     {A-OKVQA} (knowledge-based VQA), {AesBench} \citep{Huang2024AesBench} (image aesthetics), {VStar} \citep{Wu2023VStar} (spatio-temporal reasoning), {VisOnly} \citep{Kamoi2025VisOnlyQA} (vision-only recognition aggregate), {OCRBench} \citep{liu2024ocrbench} (text recognition), and {R-Bench-Dis} \citep{li2025r} (distribution-shift robustness).
    Across both Qwen2.5-VL families, reasoning-finetuned models generally underperform their base models on perception and robustness tasks, whereas {MiMo-VL-7B-RL} remains close to its SFT baseline. }
    \label{fig:radar}
    \vspace*{-5mm}
\end{figure}

\begin{wrapfigure}{r}{0.34\textwidth}
  \vspace*{-20pt} 
  \centering
  \includegraphics[width=\linewidth]{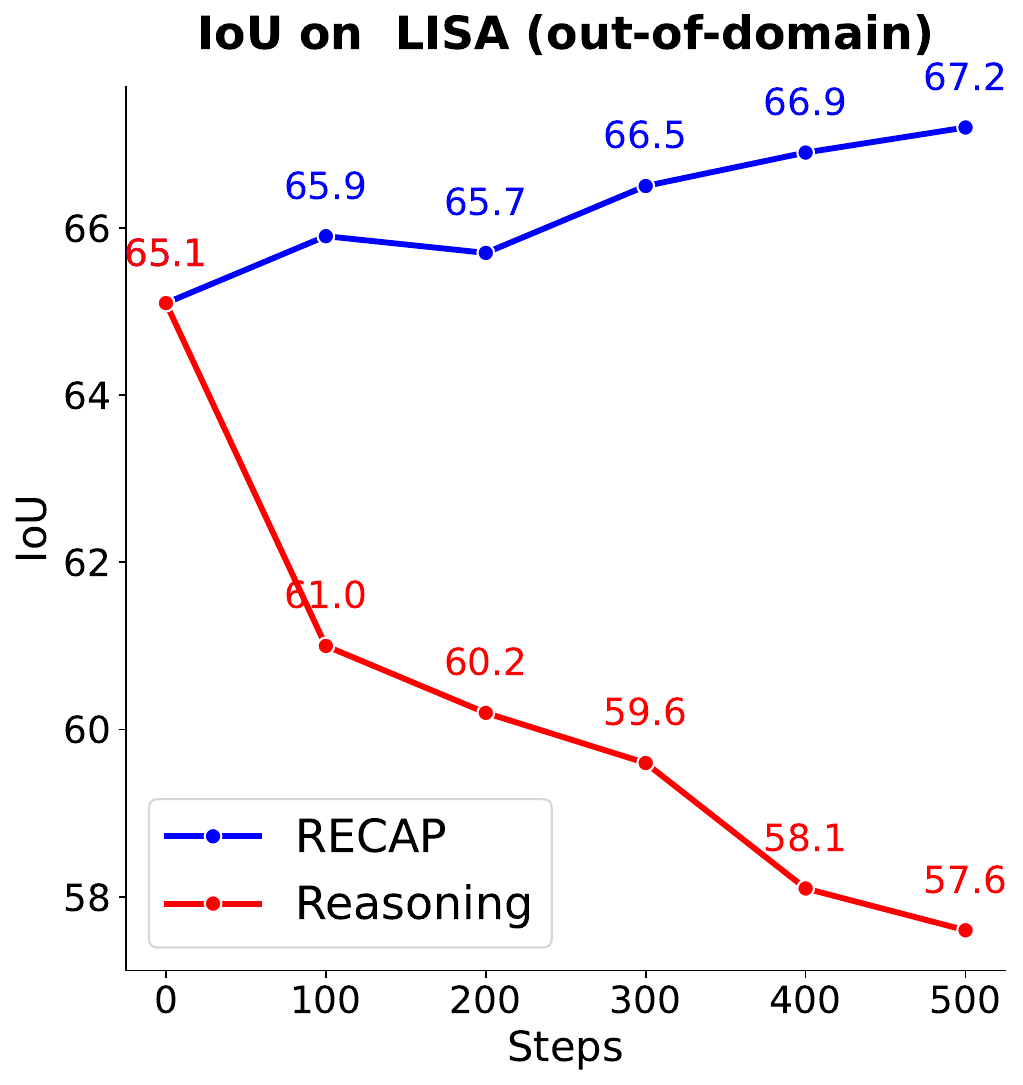}
  \caption{\textbf{Performance comparison between finetuning solely on math and  with RECAP}. Our model not only preserves base-model performance but also improves it ($\uparrow 2\%$), whereas the reasoning-only model quickly falls behind the base model after 100 iterations.
  }
  \label{fig:forgetting}
  \vspace*{-25pt}
\end{wrapfigure}
 To examine forgetting from reasoning-focused finetuning, we begin by probing the general abilities of open-source reasoning models beyond their target reasoning tasks. As shown in Fig.~\ref{fig:radar}, models finetuned for chain-of-thought or RL reasoning frequently lag behind their base counterparts on perception and robustness. For example, we observe a consistent drop on VisOnlyQA across both Qwen2.5-VL families, while MiMo-VL-7B-RL performs competitively relative to its base model on those non-reasoning tasks. We hypothesize that this is due to its specialized finetuning strategy, which employs mixed on-policy reinforcement learning to improve the model's capabilities across multiple axes beyond math and reasoning, according to the MiMo-VL-7B technical report \citep{coreteam2025mimovltechnicalreport}. However, the detailed framework and the sampling or reweighting strategy are not disclosed. These patterns support our central claim: optimizing for reasoning rewards can erode non-reasoning capabilities, motivating a continual learning method to preserve general skills during reasoning-oriented post-training. 

Our initial experiments with Qwen2.5-VL-7B show that training solely on reasoning rewards degrades performance on general capabilities; for example, performance drops by $7\%$ on the perception task, as shown in Figure \ref{fig:forgetting}. To address this, we propose gathering general-capability data and integrating it into RLVR via an online weighting mechanism. However, due to cross-domain heterogeneity, it is nontrivial to decide how much weight to assign to each loss term or reward. We then measure how different reward signals evolve during RLVR training in Section \ref{sec:method} and find that some rewards converge up to three times faster than others, suggesting that faster-converging rewards should not be emphasized later in training once the model has already learned the corresponding skills.

\begin{figure}[t]
    \centering
     \includegraphics[width=1\textwidth]{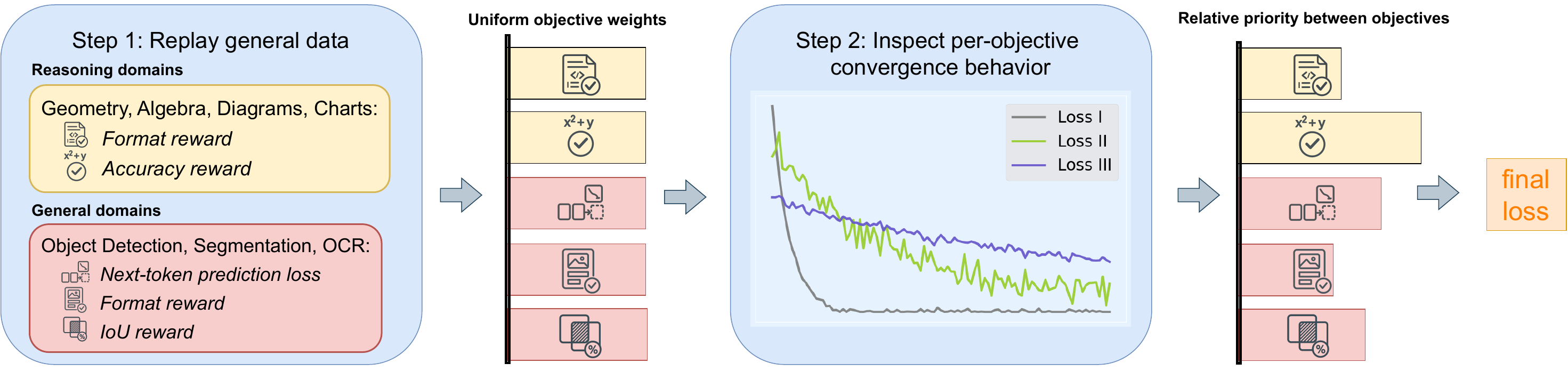}
    \caption{\textbf{Overview of RECAP.} Along with the target reasoning task, we sample data from general domains to maintain that knowledge during finetuning. Initially, the objectives of interest are weighted uniformly to optimize the main model. After a few iterations, we record the convergence behavior of individual objectives. Based on this behavior, we adjust the focus to prevent any objective from dominating and assign less weight to saturated ones. 
    }
    \label{fig:overview}
\end{figure}

Motivated by the above observations, we propose \textit{\textbf{R}eplay-\textbf{E}nhanced \textbf{CA}pability \textbf{P}reservation} (\textbf{RECAP})—a principled replay-based training strategy that mixes general data back into the RL objective and then dynamically reweights objectives based on their convergence rate and instability. As shown in Figure \ref{fig:overview}, RECAP computes the relative priorities of the objectives of interest by inspecting their convergence behavior to reweight the final loss. Experiments demonstrate that our method not only preserves general capabilities but also improves reasoning performance by allowing flexible trade-offs among reward types. Our contributions are as follows: 
\begin{itemize}
    \item We systematically re-evaluate open-source reasoning models and show that reasoning-focused finetuning consistently regresses general capabilities. This motivates replaying general-capability data during RLVR to preserve pretraining knowledge. We further show that objectives exhibit distinct convergence behavior, making commonly used, manually tuned reweighting schemes suboptimal in such scenarios.
    \item We introduce {RECAP}, a plug-in scheduler that replays general-capability data during RLVR and {dynamically} reweights both general and main-task objectives. RECAP naturally down-weights saturated format signals and refocuses capacity on harder, high-variance objectives. The method is end-to-end, magnitude-agnostic, requires no auxiliary models, and can be integrated into RLVR pipelines with negligible overhead.
    \item In our experiments, RECAP preserves or improves general capabilities while matching or exceeding the reasoning performance of reasoning-only finetuning. It consistently outperforms strong continual-finetuning baselines and is competitive with specialized open models while using less compute. Empirically, we also find that replaying general data yields shorter, more concise rationales without compromising reasoning ability.
\end{itemize}

%% file: sec/2.relatedwork.tex
\section{Related Work}
\textbf{Foundation models and post-training.}  Large transformer models pretrained on broad corpora serve as general-purpose backbones with strong abilities and wide transfer across domains \citep{brown2020language, touvron2023llama}. Post-training adapts these backbones to downstream tasks via (i) supervised finetuning, from early ULMFiT \citep{howard2018universal} to instruction tuning in FLAN \citep{wei2021finetuned} or Flan-T5 \citep{chung2024scaling}; (ii) reinforcement learning from human or AI feedback, typically combining preference modeling with policy optimization; and (iii) direct preference optimization objectives that bypass explicit reward models. For reasoning, reinforcement learning with verifiable rewards has become a common recipe: verifiers or rule-based checkers score final solutions in math and related domains, often within PPO-style pipelines \citep{guo2025deepseekr1, liu2025understanding}. Process-based neural reward models provide supervision for intermediate progress \citep{setlur2025rewarding} rather than only the final output. However, PRMs can induce reward hacking: agents learn to exploit the appearance of a correct process rather than achieving the intended outcome \citep{wang2025towards, shao2024deepseekmath}.

 \textbf{Catastrophic forgetting in continual learning and post-training.} Catastrophic forgetting describes performance regression on previously acquired skills when adapting to new data \citep{mccloskey1989catastrophic,french1999catastrophic}. Early work in this vein introduced regularization-based mitigations such as EWC \citep{kirkpatrick2017ewc}, SI \citep{zenke2017si}, and MAS \citep{aljundi2018mas} that prevent excessive changes to important parameters. Functional approaches like LwF \citep{li2016lwf} constrain outputs via distillation \citep{hinton2015kd}, while replay with small episodic memories \citep{rebuffi2017icarl,lopez2017gem,rolnick2019clear,buzzega2020der} provides consistently strong baselines across settings. 
 In RLHF post-training, a KL penalty toward the reference policy is commonly used to stabilize updates and curb over-optimization \citep{ouyang2022instructgpt}.  In addition to standard KL-regularization approaches, InstructGPT \citep{ouyang2022instructgpt} interleaves pretraining gradients with RLHF updates to reduce drift relative to KL-only regularization \citep{zheng2023secretspPO}. Concurrent works \citep{zhang2025rlep, fu2025srft} integrate verified rollouts to stabilize learning or penalize discrepancies on augmented training data \citep{wang2025perception}. However, these methods do not guarantee preservation of performance in non-target domains. Other approaches tackle forgetting by incorporating mixed, verifiable reward suites \citep{coreteam2025mimovltechnicalreport} or introducing reflection or re-attention mechanisms under RL objectives \citep{qwenlookagain2025}. In addition, recent reasoning-focused RL pipelines often reduce or remove KL to encourage exploration \citep{hu2025open,entropy2025rethink}, potentially exacerbating forgetting.

\textbf{Hybrid and multi-objective post-training.}
Recent post-training methods increasingly combine heterogeneous learning signals rather than optimizing a single reinforcement-learning objective. InstructGPT's PPO-ptx already mixed PPO updates with a pretraining language-modeling objective to mitigate capability regressions \citep{ouyang2022instructgpt}. More recent reasoning methods integrate supervised or off-policy guidance directly with online RL: SASR \citep{chen2025sasr} adaptively coordinates SFT and GRPO using gradient-norm and policy-divergence signals, SRFT \citep{fu2025srft} jointly optimizes supervised and reinforcement objectives with entropy-aware weighting, CHORD\citep{zhang2025chord} treats expert demonstrations as dynamically weighted off-policy supervision within on-policy RL, and BRIDGE \citep{chen2025bridge} formulates SFT--RL cooperation as a bilevel optimization problem. A complementary line studies multiple rewards or domains within RL. MiMo-VL  \citep{coreteam2025mimovltechnicalreport} jointly optimizes RLVR and RLHF signals with rule- and model-based rewards, while MoDoMoDo \citep{liang2025modomodomultidomaindatamixtures} learns data mixtures for multi-domain multimodal RLVR using a surrogate of downstream training outcomes. More directly, multi-objective RL methods  \citep{delangis2024dynamic} dynamically rebalance rewards using discriminator-gradient magnitudes, weak-objective performance \citep{chen2025drbo} or Pareto-oriented hypervolume and gradient criteria \citep{lu2025dynamicreward}. 

%% file: sec/3.background.tex
\section{Background}


This section first provides the essential background on standard post-training for large language models, covering (i) supervised finetuning (SFT), (ii) RL-based alignment from preferences, (iii) reinforcement learning with verifiable rewards, and (iv) GRPO for long-form reasoning. Our approach builds on recent progress in reasoning-centric LLMs—exemplified by DeepSeek-R1 \citep{guo2025deepseekr1} and other contemporary models \citep{ji2025thinking, yu2025perception, chen2025seed15thinking, wang2025vl}. We adopt GRPO as our primary RL algorithm because it effectively reduces memory and compute overhead relative to standard PPO.

 \textbf{Supervised finetuning (SFT).} Let an LLM with parameters $\theta$ induce a conditional policy $\pi_\theta(\cdot\mid x)$ over responses $y$ to a prompt $x$. SFT optimizes the negative log-likelihood on instruction–response pairs $\mathcal D=\{(x^{(i)},y^{(i)})\}_{i=1}^N$:

$$
\mathcal L_{\mathrm{SFT}}(\theta)
= -\sum_{i=1}^{N}\log \pi_\theta\!\big(y^{(i)}\!\mid x^{(i)}\big).
$$

SFT has been central to transferring general-purpose LMs to downstream instruction following and broad zero-shot generalization (e.g., ULMFiT \citep{howard2018universal}; instruction-tuned FLAN \citep{wei2021finetuned}), and it typically provides the initialization for subsequent preference- or reward-based alignment.

\textbf{RL-based post-training.} Reinforcement learning from human feedback (RLHF) fits a reward model $r_\phi(x,y)$ from pairwise human preferences \citep{ziegler2019fine, rafailov2023direct, lambert2025reinforcement}, commonly using a Bradley–Terry likelihood \citep{bradley1952rank}, and then maximizes reward while regularizing toward $\pi_{\text{ref}}$ (often with a KL penalty) via policy optimization such as PPO \citep{schulman2017proximal}:

$$
\max_\theta\ \mathbb E_{x\sim\mu,\ y\sim\pi_\theta(\cdot\mid x)}\!\Big[r_\phi(x,y)\Big]
\ -\ \beta\,\mathbb E_{x\sim\mu}\!\Big[D_{\mathrm{KL}}\big(\pi_\theta(\cdot\mid x)\,\|\,\pi_{\text{ref}}(\cdot\mid x)\big)\Big].
$$

This pipeline improves helpfulness and harmlessness while retaining base-model competence; see early LM-preference work and InstructGPT \citep{ouyang2022instructgpt} for canonical formulations, and PPO for the underlying stable policy-gradient updates \citep{stiennon2020learning}.

\textbf{Reinforcement learning with verifiable rewards (RLVR).} In settings with programmatic or automatic verifiers (e.g., exact-match answers, execution-based checks, or constraint checkers), RLVR replaces learned human-preference rewards with verifiable signals $r(x,y)\in[0,1]$. This reduces labeler noise and can better target reasoning fidelity by rewarding demonstrably correct steps or outcomes, while retaining the same KL-regularized RL form \citep{lightman2023let}.

\textbf{GRPO for long-form reasoning.} Group Relative Policy Optimization (GRPO) is a PPO-style algorithm tailored for LLM reasoning that forgoes a learned critic and instead computes advantages from group-normalized sequence rewards. For each prompt $x$, sample a group of $G$ rollouts $O=\{o_i\}_{i=1}^G$ from a frozen rollout policy $\pi_{\theta_{\text{old}}}$. Let $R_i$ be the verifiable sequence-level reward (e.g., exact-match correctness), and define the group-normalized advantage $\widehat A_i=\big(R_i-\mathrm{mean}(R)\big)/\mathrm{std}(R)$. With token-wise importance ratio

$$
r_{i,t}(\theta)=\frac{\pi_\theta(o_{i,t}\mid x,o_{i,<t})}{\pi_{\theta_{\text{old}}}(o_{i,t}\mid x,o_{i,<t})},
$$

GRPO maximizes the clipped surrogate plus KL regularization:

$$
\mathcal J_{\text{GRPO}}(\theta)=\mathbb E\!\left[
\frac{1}{G}\sum_{i=1}^{G}\frac{1}{|o_i|}\sum_{t=1}^{|o_i|}
\min\!\Big\{r_{i,t}(\theta)\widehat A_i,\ \mathrm{clip}\!\big(r_{i,t}(\theta),1-\epsilon,1+\epsilon\big)\widehat A_i\Big\}
\ -\ \beta\,D_{\mathrm{KL}}\!\big(\pi_\theta\ \|\ \pi_{\text{ref}}\big)
\right].
$$
By normalizing across a group of responses for each prompt, GRPO stabilizes updates without a critic, which is preferable when long chain-of-thought rewards are sparse and verifier-based. Empirically, GRPO has been shown to boost mathematical-reasoning performance in open models \citep{shao2024deepseekmath}.

\textbf{General ability degradation.} Unlike traditional continual-learning studies, modern post-training pipelines must jointly consider gains in reasoning and retention of general abilities acquired during pretraining (e.g., perception, grounding, instruction following, safety). Let $\mathcal{G}=\{G_1,\dots,G_M\}$ denote a suite of general-ability tasks and $\mathcal{R}=\{R_1,\dots,R_L\}$ a suite of reasoning datasets.

%% file: sec/4.method.tex
\section{Our Method: Replay-Enhanced CApability Preservation (RECAP)}
\label{sec:method}

We address forgetting in RLVR by (i) replaying general-capability data alongside reasoning data, and (ii) dynamically reweighting objectives online using local estimates of progress and instability for individual objectives. Below, we present how RECAP governs loss coefficients and shifts the optimization away from saturated objectives toward underperforming or volatile ones—without changing the underlying RL algorithm.

\textbf{Setting.} In the context of supervised learning,  let $\mathcal{D}=\{\mathcal{D}_n\}_{n=1}^{N}$ be $N$ domains and
$\ell_{n,k}^{(t)}(\theta)$ the mini-batch loss of objective
$k\in\{1,\dots,K\}$ on domain $n$ at iteration $t$ for parameters
$\theta$. Note that $K \geq N$ because some tasks use more than one reward or objective.
The model parameters $\theta$ are thus optimized by minimizing the average loss across objectives:
\begin{equation*}
L_k^{(t)}\;=\;\frac{1}{N}\sum_{n=1}^{N}\ell_{n,k}^{(t)}(\theta).
\end{equation*}
Our framework acts on $\{L_k^{(t)}\}_{k=1}^K$ regardless of whether each $L_k$ arises from an RL reward surrogate or a supervised learning loss term.

\textbf{Per-objective convergence rate and stability of convergence.} Due to the unstable nature of RL training, we cannot rely solely on the per-step objective or reward value to compute the reweighting coefficients. Instead, for each objective $k$, we measure the \emph{convergence rate} over a sliding window of length $2 \times W$ by computing the current window average and the previous window average:
\begin{align*}
\underbrace{\mu_k^{(t)} = \frac{1}{W}\sum_{s=t-W+1}^{t}L_k^{(s)}}_{\text{estimated current loss value}}&, \qquad\qquad\qquad\qquad
\underbrace{\tilde{\mu}_k^{(t)} = \frac{1}{W}\sum_{s=t-2W+1}^{t-W}L_k^{(s)}}_{\text{estimated old loss value}}
\end{align*}

and the \emph{instability} (coefficient of variation) in the same window:
\begin{align*}
\sigma_k^{(t)} =&
      \sqrt{\frac{1}{2W-1}\sum_{s=t-2W+1}^{t}\bigl(L_k^{(s)}-\mu_k^{(t)}\bigr)^2}.
\end{align*}
Based on these measures, we form two signals: (i) the \fbox{\emph{convergence rate}
$c_k^{(t)}=\tilde{\mu}_k^{(t)}/{\mu}_k^{(t)}$}
captures how quickly the loss improves, while (ii) the \fbox{\emph{inverse signal-to-noise ratio}
$i_k^{(t)}=\sigma_k^{(t)}/(\mu_k^{(t)}+\tilde{\mu}_k^{(t)})$} captures loss instability. Intuitively, $c_k^{(t)}>1$ indicates recent improvement (loss dropping relative to the previous window), while $c_k^{(t)}\!\approx\!1$ signals saturation. The term $i_k^{(t)}$ is larger when the objective is noisy or unstable.

\textbf{Relative priority between domains.}
We convert these signals into normalized coefficients via a temperature-controlled softmax. Given temperature $T>0$, we define the priority score of the $k$-th objective as $s_k^{(t)}$ and compute the coefficients $\lambda$ for reweighting objectives.
\begin{equation}
    s_k^{(t)} \;=\; c_k^{(t)} +\, i_k^{(t)},\qquad
\lambda_k^{(t)} \;=\;
\frac{K\,\exp\!\bigl(s_k^{(t)}/T\bigr)}
     {\sum_{i=1}^{K}\exp\!\bigl(s_i^{(t)}/T\bigr)}. \label{eq:main}
\end{equation}
The prefactor $K$ preserves average scale so that $\frac{1}{K}\sum_k \lambda_k^{(t)}\!=\!1$. Lower $T$ sharpens priorities while higher $T$ approaches uniform mixing. We set $T\!=\!5$ by default in our experiments. 

\textbf{Overall training objective.}
At step $t$, we minimize the following weighted objective:
\[
\mathcal{L}^{(t)}(\theta)
\;=\;
\frac{1}{K}\sum_{k=1}^{K}\lambda_k^{(t)}\,L_k^{(t)}.
\]
Optimizing $\theta$ with \(\nabla_\theta \mathcal{L}^{(t)}\) steers
learning toward objectives that are \emph{both} slow to converge
(high $c_k$) \emph{and} fluctuating (high $i_k$), while leaving
well-learned, stable objectives with lower weight.
The scheme reduces to the standard equal
weighting as \(T\to\infty\) (i.e., uniformly sample from all domains without loss reweighting).

\clearpage

\begin{figure}[t]
    \centering
     \includegraphics[width=1\textwidth]{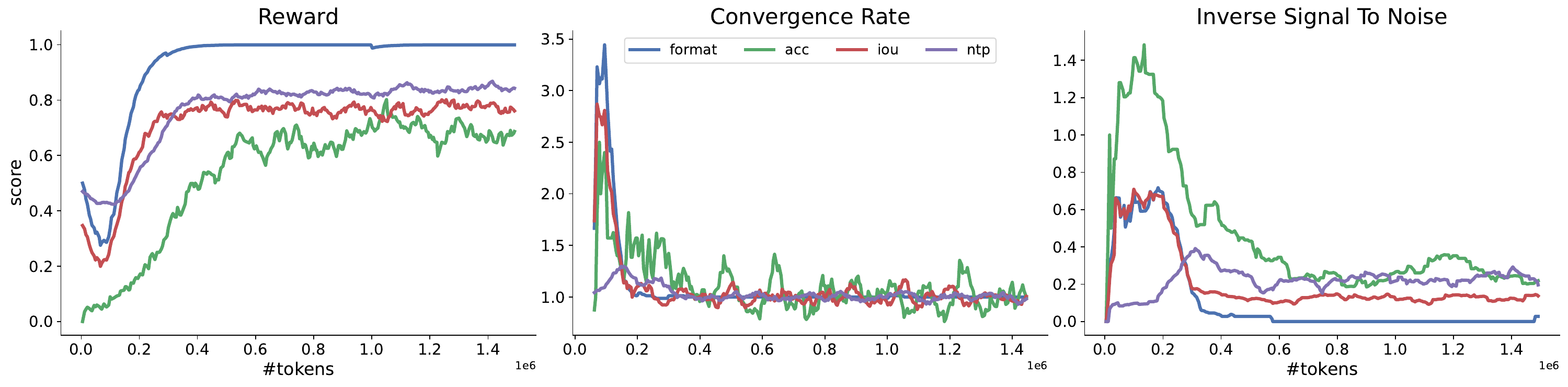}
    \caption{\textbf{Different rewards exhibit different convergence behavior.} While the \textcolor{indigoblue}{format} reward is easy to optimize and initially has the highest convergence rate, it quickly saturates and thus yields a near-unity convergence rate ($c\sim1$) and low instability ($i\sim0$) after 50 steps. By contrast, the reasoning \textcolor{forestgreen}{accuracy} fluctuates the most, thereby steering the optimization toward the corresponding objective. \textcolor{magentaaccent}{IoU} and \textcolor{violetpurple}{ntp} denote the IoU reward and next-token-prediction accuracy during training. The result is obtained in the first setting in our experiments below. }
    \label{fig:reward}
\end{figure}

Figure \ref{fig:reward} illustrates the insight behind our proposed method, in which we finetune the model on a reasoning dataset (tracked by \textcolor{forestgreen}{accuracy}) while replaying perception data (\textcolor{magentaaccent}{IoU}) and an SFT dataset for general-capability retention. Early in training, the \textcolor{indigoblue}{format} signal is easy to optimize and saturates quickly, so its $c$ falls toward $1$ and $i$ toward $0$, reducing its priority. After 100 steps, many signals plateau (convergence rate $c\!\approx\!1$) while still differing in stability. Among them, the reasoning reward fluctuates the most ($\sim 0.3$), yielding higher $i$ and thus higher $\lambda$. At this point, the model has learned to answer according to the predefined template; thus, the corresponding signal-to-noise ratio of the formatting reward is $\approx 0$. This motivates combining \emph{both} progress ($c$) and instability ($i$) in Equation \ref{eq:main}, as they complement each other. Tuning the trade-off between these two terms offers finer-grained control and can improve performance. However, for simplicity, we take their unweighted sum $s=c + i$, which performs consistently well in our experiments.

\paragraph{Entropy-regularized interpretation of RECAP reweighting.}
We can show that the RECAP reweighting rule arises as the closed-form solution of an entropy-regularized priority allocation problem. At iteration \(t\), let:
\[
s^{(t)} = \big(s_1^{(t)}, \ldots, s_K^{(t)}\big) \in \mathbb{R}^K
\]
denote the per-objective priority scores. In RECAP, we define
$s_k^{(t)} = c_k^{(t)} + i_k^{(t)},$
where \(c_k^{(t)}\) measures the recent convergence behavior of objective \(k\), and \(i_k^{(t)}\) captures its instability.

Rather than manually assigning weights to those objectives, we
interpret the weights as an allocation vector:
\[
p \in \Delta^K := \left\{p \in \mathbb{R}_{\geq 0}^K : \sum_{k=1}^K p_k = 1 \right\}.
\]
Given a temperature parameter \(T > 0\), the entropy-regularized priority allocation objective has the following formula:
\[\quad\quad\quad\quad\quad\quad\quad\quad\quad\quad\quad\quad\quad\quad\quad\quad\quad\quad
\Phi_t(p)
=
\langle p, s^{(t)} \rangle 
+
T \mathcal{H}(p),
\;\quad\quad \text{where}\quad
\mathcal{H}(p)
=
-\sum_{k=1}^K p_k \log p_k,
\]
We consider the entropy-regularized priority allocation problem:
\[
\max_{p \in \Delta^K} \Phi_t(p).
\label{eq:entropy-priority-allocation}
\]
The linear term assigns higher mass to objectives with larger RECAP priority scores, while the entropy term prevents degenerate allocation to a single objective and encourages smoother mixtures. The temperature \(T\) controls this trade-off: smaller \(T\) yields sharper allocations, whereas larger \(T\) approaches uniform weighting. The following result characterizes the solution.

\begin{theorem}[Entropy-regularized priority allocation]
For any \(T > 0\) and any score vector \(s^{(t)} \in \mathbb{R}^K\), this optimization problem has a unique maximizer \(p^{(t),\star} \in \Delta^K\), given by
\[
p_k^{(t),\star}
=
\frac{\exp\left(s_k^{(t)}/T\right)}
{\sum_{j=1}^K \exp\left(s_j^{(t)}/T\right)},
\qquad
k = 1, \ldots, K.
\]
\end{theorem}

The proof is deferred to Appendix~\ref{app:entropy-recap}. This theorem shows that the softmax form used in RECAP is not merely a heuristic normalization, it is  an entropy-regularized allocation problem over objectives. 

%% file: sec/5.experiment.tex
\section{Experiments}
\label{sec:experiment}
\subsection{Experimental Settings}

To evaluate our proposed approach, we conduct experiments on two base models, Qwen2.5-VL-3B and Qwen2.5-VL-7B, across two complementary setups at different scales.
\textbf{RLVR-Only Setting:} This smaller setup follows the experimental configuration of \citet{liang2025modomodomultidomaindatamixtures}, focusing on domain-specific RLVR, which serves as an \emph{upper-bound} baseline for static data-mixture approaches and as our closest baseline.
 The Qwen2.5-VL-3B model is trained until data from a particular domain is exhausted. 
 Since \citet{liang2025modomodomultidomaindatamixtures} enforces binary (0–1) rewards across all domains and thus cannot directly handle non-RL tasks, we further extend our evaluation to a \textbf{Hybrid Setting:} To bridge RL and supervised training paradigms, we finetune Qwen2.5-VL-7B under a larger mixed-objective regime that combines RLVR with SFT-style training. Specifically, we use ThinkLite-VL-70k \citep{wang2025sota} while jointly replaying perception-oriented datasets such as RefCOCO \citep{kazemzadeh-etal-2014-referitgame} and LLaVA-OneVision OCR \citep{li2024llava}.
 To isolate the effect of replay and dynamic reweighting and for ease of implementation, we uniformly sample across data sources by default and reweight only the objectives of interest. Unless otherwise noted, we disable the reference-KL penalty to disentangle the effectiveness of regularization approaches \citep{li2017learning} from that of our replay mechanism. We also include a comparison with this regularization-based approach in our list of established baselines for continual learning below:

\begin{itemize}
    \item \textbf{Reasoning-only:} We train only on the target reasoning task with fixed reward weights and no replay. This baseline represents standard task-specific RLVR finetuning without any explicit mechanism for preserving general capabilities.
    \item \textbf{PropMix:} We include general-capability data during finetuning and sample domains in proportion to their source-dataset sizes. Objective losses are not reweighted.
    \item \textbf{Uniform:} We sample data uniformly across domains while keeping all objective weights fixed.
    \item \textbf{Coreset:} We replay a size-limited subset of general-capability data, set to half of the reasoning-data volume in our setup, following standard coreset-style replay methods \citep{rebuffi2017icarl,chaudhry2019agem}.
    \item \textbf{LwF:} We sample data uniformly across domains and add KL regularization with coefficient $\beta=0.01$. We refer to this variant as LwF because it follows the principle of Learning without Forgetting \citep{li2017learning} by constraining the updated model to remain close to a reference policy.
\end{itemize}

For context, we also include representative open-source vision-language models specializing in reasoning that are derived from the corresponding base models in each experiment. We list them here for easier benchmarking and do not aim to outperform them, as those models often undergo many complex training pipelines. Models are evaluated with LMMS-Eval \citep{zhang2024lmmsevalrealitycheckevaluation}.
\subsection{Experimental Results}


According to Table \ref{tab:small}, RL training on the reasoning domain consistently improves base-model performance on both reasoning and perception benchmarks. On SCIENCEQA in particular, RL lifts the Qwen2.5-VL-3B score from $6$ to $60$. On this benchmark, our proposed method even outperforms the comparison open-source reasoning models. We consider MoDoMoDo the \textit{upper-bound} approach among static data-mixture approaches because (i) it has access to target-task performance during finetuning, which requires rerunning experiments if new target tasks are introduced, and (ii) it trains multiple proxy models of the same size as the baseline models to learn test performance as a function of the mixing ratio, which is computationally expensive,

\begin{table}[!ht]
\centering
\caption{\textbf{Benchmark results in the RLVR-only setting.} We report accuracy scores on six benchmarks, where the MoDoMoDo baseline is trained to maximize performance. For this table only, we use the rule-based evaluator on the MathVista dataset instead of "gpt-3.5-turbo" to align with \citet{liang2025modomodomultidomaindatamixtures}.
}
\label{tab:small}
\resizebox{.95\textwidth}{!}{
\begin{tabular}{lcccccc}
\toprule
\textbf{Model} & \textbf{SAT} & \textbf{ScienceQA} & \textbf{MathVista (mini)} & \textbf{ChartQA} & \textbf{InfoVQA} & \textbf{MMMU} \\
\midrule
\rowcolor{lightgray} \multicolumn{7}{l}{\emph{Open-source reasoning baselines}} \\
VLAA-Thinker-3B         & 49.38    & 14.63        & 30.4  & 45.84 & 30.81 & 32.22 \\
MM-R1-MGT-PerceReason             & 50.83       & 34.21       & 33.4  & 44.88 & 61.42     & 40.22    \\
Ocean\_R1\_3B\_Instruct           & 59.49       & 68.72       & 38.7  & 54.00 & 38.02 & 40.89 \\
Qwen2.5VL-3b-RLCS                 & 24.12       & 21.32       & 17.2  & 3.32      & 10.86      & 27.11    \\
vision-grpo-qwen-2.5-vl-3b        &    50.57         &    4.17        &   32.4 & 67.80 & 58.29 &   37.22    \\
Qwen2.5-VL-3B-Instruct-GRPO-deepmath        &    34.70         & 45.27           &   32.3 & 70.24 & 49.75 &   39.11    \\

\midrule
\rowcolor{lightgray} \multicolumn{7}{l}{\emph{Qwen2.5-VL-3B and our variants}} \\
Base model           & 43.98  &   6.20     & 23.6  & 43.88 & 32.02 & 38.67      \\
Uniform                             & 44.55  & 64.85  & 32.4  & 69.68 & 58.30 & 39.44 \\
MoDoMoDo                          & 49.95  & 65.74  & 32.2  & \textbf{70.40} & 59.88 & 39.11  \\
\hline
 \rowcolor{carolinablue!60} \textbf{RECAP} & \textbf{55.19} & \textbf{71.59}  & \textbf{33.2}  & \textbf{70.40} & \textbf{60.78} & \textbf{42.44} \\
\bottomrule
\end{tabular}
}
\end{table}

especially in the context of reinforcement learning. Even after selecting an “optimal” mixture, the method still depends on hand-tuned reward weights (e.g., doubling accuracy and IoU relative to formatting rewards). These trade-off coefficients are also set differently in prior work without clarification, which limits generality.


Table \ref{tab:large} reports the performance of models finetuned from Qwen2.5-VL-7B across multiple benchmarks in the hybrid setting, showcasing a more general scenario than RLVR-only training. Overall, RECAP achieves the best or runner-up performance across datasets, except for VizWiz, where it still improves base-model performance by more than $10\%$. Compared with the base model, naively finetuning on the reasoning domain causes significant forgetting, especially on tasks that do not require extensive thinking. For example, finetuning on ThinkLite-VL-70k reduces segmentation ability from 65.13 to 57.58 on LISA. Meanwhile, replaying general data helps preserve performance across all baselines on this dataset ($\approx 64$). Compared with uniform sampling, LwF achieves similar scene-understanding performance while obtaining lower scores on the reasoning benchmark (e.g., 29.59 vs 31.91 on MMMU-PRO). Similar behavior is observed in \citep{wang2025perception, hu2025open}, where this term is removed for more plasticity. Among all baselines, our proposed method obtains the highest segmentation score, boosting base-model performance by more than $2\%$. This improvement highlights the impact of loss reweighting over the uniform baseline.

\begin{table}[!ht]
\caption{\textbf{Benchmark results in large hybrid setting.}
We report accuracy scores (higher is better) on nine perception and reasoning benchmarks.
Rows above the break are open-source reasoning models with different backbones; the lower block compares variants {finetuned from the same Qwen2.5-VL-7B base model}.
Bold = best; \underline{underline} = second best within the Qwen2.5-VL-7B family.}
\label{tab:large}
\resizebox{\textwidth}{!}{
\centering
\begin{tabular}{lccccccccc}
\toprule
\textbf{Model} & \textbf{LISA} & \textbf{MMMU-PRO} & \textbf{AI2D} & \textbf{MathVista} & \textbf{MathVision} & \textbf{MathVerse} & \textbf{MMBench} & \textbf{VizWiz} & \textbf{OCRBenchv2} \\
\midrule
\rowcolor{lightgray} \multicolumn{10}{l}{\emph{Open-source reasoning baselines}} \\
VLAA-Thinker-7B & 63.14 & 26.30   & 75.45    & 63.90 & 11.18 & 29.87    & 75.95 & 47.57  & 40.23 \\
Vision-R1-7B               &  47.30 & 26.76   & 0.00    & 61.80 & 18.75 & 23.32    & 69.46    & 53.12   & 24.63    \\
OpenVLThinker-7B           & 42.73 & 21.79   & 59.94   & 59.10 & 5.59    & 19.26    & 71.53    &52.89  & 28.30   \\
\midrule
\rowcolor{lightgray} \multicolumn{10}{l}{\emph{Qwen2.5-VL-7B and our variants}} \\
Base model      & 65.13 & 25.55   & 67.62    & 61.70 &  9.54 & 26.29 & 71.82 & 50.82  & 39.49  \\
LwF                        & 65.08 & 29.59   & 73.93 & 63.90 & 18.42 & 33.98 & 73.11 & 53.12  & \underline{39.56}     \\
PropMix                    & \underline{66.80} & 31.39   & 75.32 & 63.40 & 21.05 & 34.75 & 73.54 & 57.05 & 37.60   \\
Uniform                    & 65.18 & 31.91   & 76.43    & 65.60 & 22.13   & 36.07    & 75.34 & 54.05 & 38.06   \\
Coreset                   & 64.82 & 31.91   & \textbf{79.92} & \textbf{66.90} & 23.36 & 37.58    & \underline{78.09} & \textbf{63.76} & 35.49   \\
Reasoning-only             & 57.58 & \underline{33.87}   & 74.97 & 65.50 & 24.87    & \underline{40.74} & 77.84 &  \underline{62.45}   & 38.55   \\
\hline
\rowcolor{carolinablue!60} \textbf{RECAP}                       & \textbf{67.24} & \textbf{34.15}   & \underline{78.21} & \underline{66.70}    & \textbf{25.11}    & \textbf{40.83}    & \textbf{78.52}    & 61.97   & \textbf{39.72}    \\

\bottomrule
\end{tabular}
}
\end{table}



\subsection{Ablation Studies}
\begin{figure}[!ht]
    \centering
     \includegraphics[width=.9\textwidth]{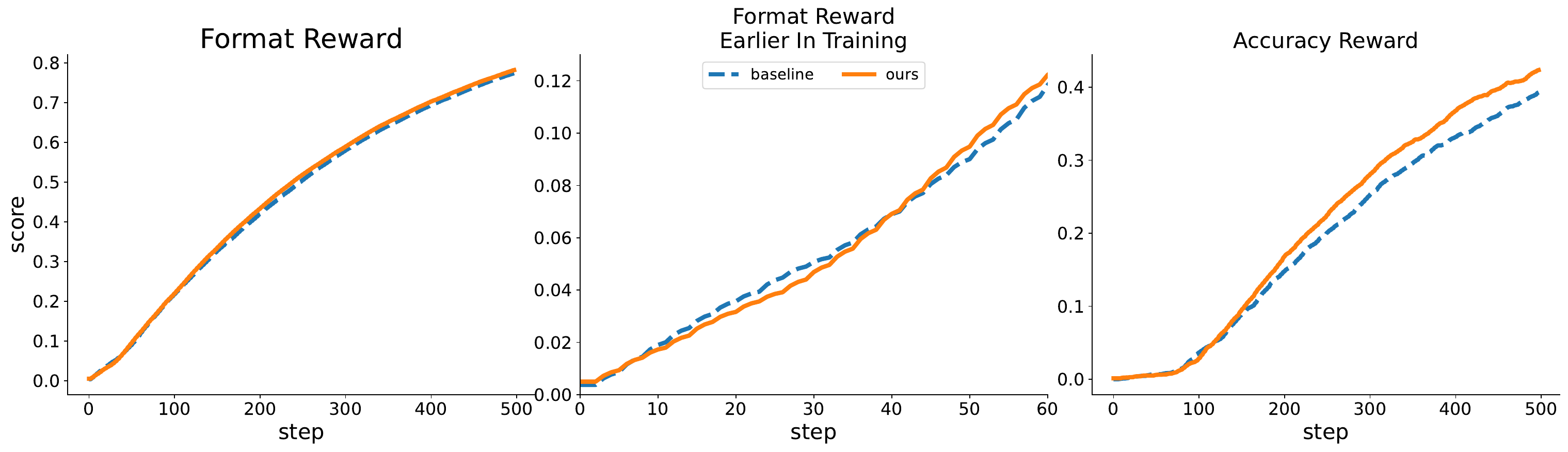}
    \caption{\textbf{Evolution of format and accuracy rewards on the reasoning domain during training:} Training curves for the format reward over full training (left), an early-training zoom (middle), and the accuracy reward (right). While the uniform baseline is better at maximizing the format reward, it falls behind our proposed method later in terms of accuracy, as we prioritize correct solutions over formatting once the model can follow the predefined template. Curves are smoothed with an exponential moving average for readability.
    }
    \label{fig:ablation}
\end{figure}

We conduct an ablation study on the large hybrid setting by comparing the accuracy and formatting rewards of our proposed method with the uniform baseline to isolate the effect of our reweighting. The uniform baseline employs identical hyperparameters, including the data-sampling and model-training pipelines, yet differs from ours only in the loss reweighting mechanism ($\lambda_k=1/K$). In Figure \ref{fig:ablation}, we present the curves for the format and accuracy rewards during training. In the early phase, the baseline's format reward increases faster—consistent with format being a low-variance, easy-to-optimize signal—yet a crossover soon appears and our method surpasses it as training progresses (step 40). In contrast, for the accuracy reward, our method opens a growing lead over time (right). This behavior aligns with our scheduler: once the format objective shows fast convergence and low instability, its weight is down-regulated and capacity is reallocated to harder, higher-variance objectives (e.g., accuracy), avoiding over-optimization of formatting while improving task correctness.

We also empirically find that using the same format reward for different domains is suboptimal. We start by examining the approach from \citet{liang2025modomodomultidomaindatamixtures}, which employs the same thinking reward on every domain, including scene-understanding tasks. In Figure \ref{fig:length}, we plot the response length on the segmentation task during

  \begin{minipage}[b]{0.49\textwidth}
    \vspace*{0mm}
    \centering
\includegraphics[width=1\columnwidth]{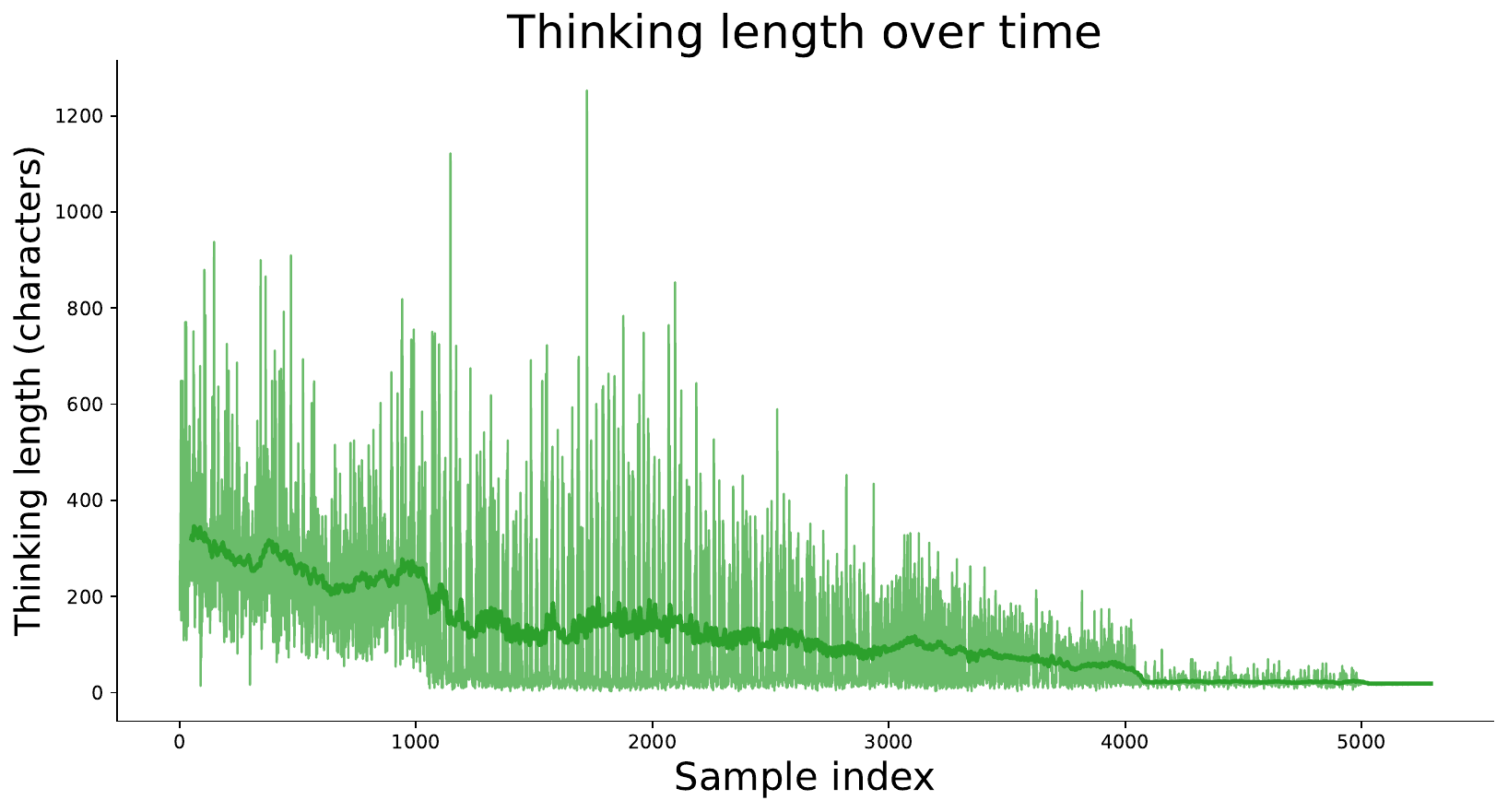}
    \vspace*{-0mm}
    \captionof{figure}{\textbf{Thinking length on a segmentation task during finetuning.} When a uniform “thinking reward’’ is applied to all domains, the model quickly learns that long chains of thought are unnecessary for segmentation. The average response length drops from several hundred characters at the start of training to tens (often near zero) later on. 
    \label{fig:length}
    }
    \vspace*{0mm}
  \end{minipage}
  \hfill
  \begin{minipage}[b]{0.49\textwidth}
    \centering
    {\vspace*{0mm}
\includegraphics[width=1\columnwidth]{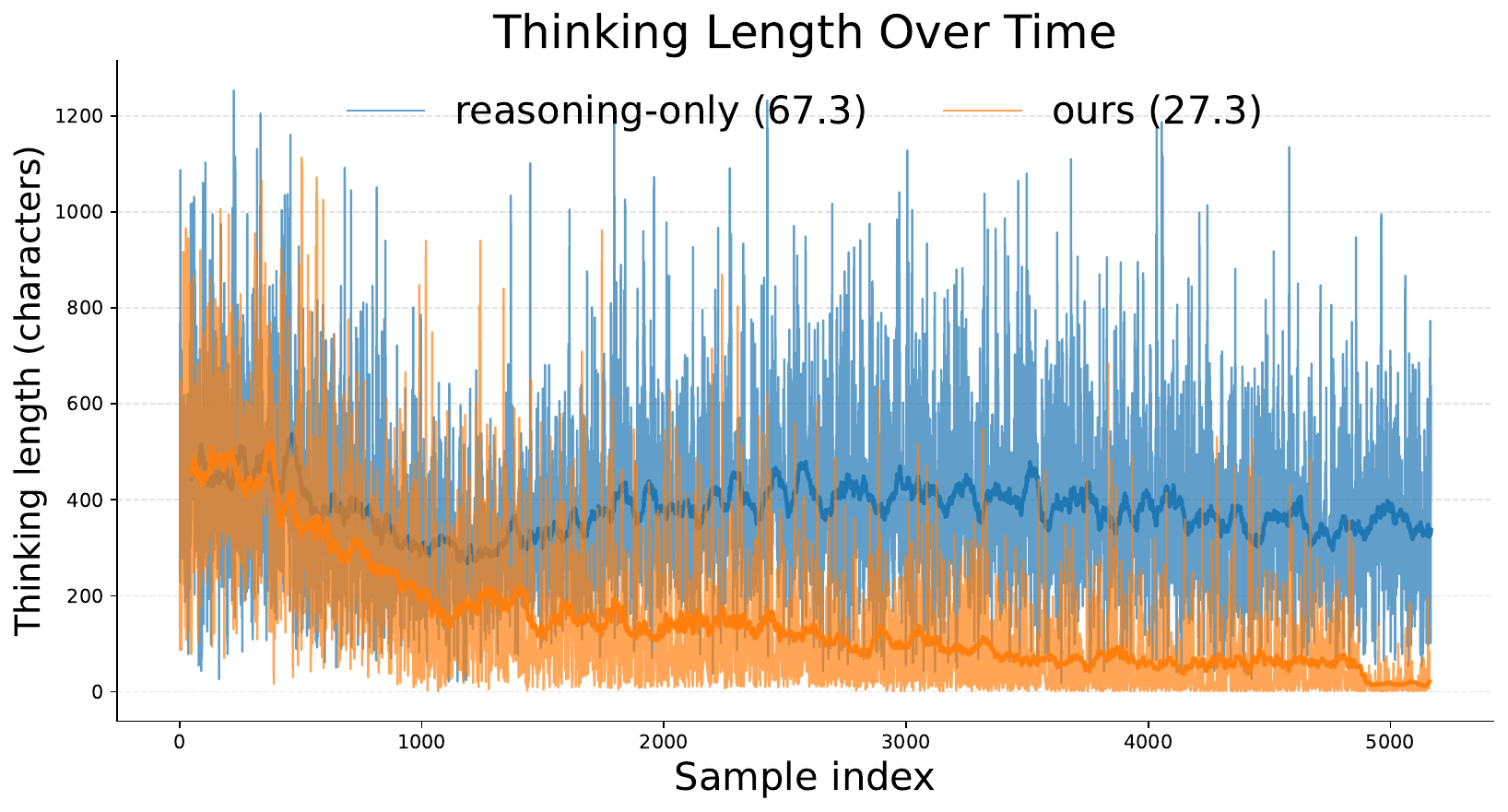}
    \vspace*{-0mm}}
    \captionof{figure}{\label{fig:think_length}\textbf{Thinking length on a reasoning task during finetuning.}
We compare a model trained {reasoning-only} (\textcolor{blue}{blue}) against our RECAP method (\textcolor{orange}{orange}). Highlighted curves show the running average thinking length per example, where our method generates only $\sim 27.3$ words per question, compared with the baseline ($\sim 67.3$). 
}
    \vspace*{0mm}
  \end{minipage}

training and find that the Qwen2.5-VL model rapidly trims its chain-of-thought and answers the question directly later in training. This behavior suggests that explicit reasoning is unnecessary for such perception tasks and that encouraging long rationales can even be detrimental. We also include qualitative examples in the appendix to show how the model gradually suppresses its reasoning trace during training. Motivated by this observation, in our broader setting, we keep {answer-format} rewards for perception domains (no thinking) and reserve thinking rewards for tasks that truly benefit from step-by-step reasoning.

Figure \ref{fig:think_length} tracks the length of the generated {thinking} segment on the reasoning task throughout training. When trained only on reasoning data, the model maintains long chains of thought with high variability. In contrast, mixing general-capability replay with dynamic objective reweighting progressively reduces thinking length and stabilizes variance, converging to concise rationales (60\% reduction, $67\rightarrow27$ words on average) while preserving accuracy. This shorter reasoning directly improves inference efficiency—fewer generated tokens reduce latency and compute cost—without sacrificing problem-solving quality.

%% file: sec/6.conclusion.tex
\section{Conclusion}
In this paper, we investigate forgetting in recent reasoning-focused vision-language models and find that these models exhibit clear forgetting of general knowledge obtained during pretraining. Motivated by this, we propose an approach that replays general data during finetuning and uses a plug-in method to reweight objectives without incurring the additional cost of training external models. On reasoning benchmarks, our proposed method not only preserves general knowledge but also improves target reasoning performance by properly reweighting the rewards for those tasks.

\section*{Limitations}
Our framework RECAP is generic and extends naturally beyond RLVR and SFT to preference- and alignment-based objectives \citep{rafailov2023direct,garg2025ipo,hong2024orpo,ethayarajh2024kto} and process reward models \citep{lightman2023verify,setlur2024rewarding}. However, due to constraints on the training datasets available for this work, our empirical evaluation focuses on RLVR and standard SFT settings. We expect RECAP to yield similar gains over uniform or manually tuned baselines with any heterogeneous objective sets, but we leave a comprehensive evaluation across non-RL objectives to future work. In practice, applying our scheduler to non-RL losses does not require an expensive search over coefficients or per-objective normalization due to its magnitude-agnostic nature.
\section*{Acknowledgments}
We would like to thank Riham Mansour and Misha Bilenko for their steadfast leadership and support. We would also like to thank the Meta legal and policy team for their review and approval of this research.


%% file: sec/appendix.tex
\section{Experimental setup}

In this section, we provide detailed statistics of the training datasets used in our experiments, along with implementation details and additional experimental results. 

\subsection{Proof of Entropy-Regularized Priority Allocation}
\label{app:entropy-recap}

We provide the proof that the RECAP reweighting rule is the unique solution of an entropy-regularized priority allocation problem.

\begin{lemma}[Existence of an optimizer]
\label{lem:entropy-existence}
For any \(T > 0\) and any score vector \(s^{(t)} \in \mathbb{R}^K\), the optimization problem in Equation \ref{eq:entropy-priority-allocation} admits at least one maximizer \(p^\star \in \Delta^K\).
\end{lemma}

\begin{proof}
The feasible set \(\Delta^K\) is compact. The map
\[
p \mapsto \langle p, s^{(t)} \rangle
\]
is continuous, and the entropy \(\mathcal{H}(p)\) is continuous on \(\Delta^K\) under the convention \(0 \log 0 = 0\). Therefore, \(\Phi_t\) is continuous on a compact set. By the extreme value theorem, \(\Phi_t\) attains its maximum on \(\Delta^K\). Hence, there exists at least one maximizer \(p^\star \in \Delta^K\).
\end{proof}

\begin{lemma}[Interior optimality]
\label{lem:entropy-interior}
Any maximizer \(p^\star\) of Equation \ref{eq:entropy-priority-allocation} lies in the interior of the simplex, i.e.,
\[
p_k^\star > 0,
\qquad
k = 1, \ldots, K.
\]
\end{lemma}

\begin{proof}
Assume for contradiction that \(p^\star\) is a maximizer and that \(p_j^\star = 0\) for some \(j\). Since \(\sum_{k=1}^K p_k^\star = 1\), there exists some index \(i \neq j\) such that \(p_i^\star > 0\).

For \(\varepsilon \in (0, p_i^\star)\), define the feasible perturbation
\[
p(\varepsilon)
=
p^\star + \varepsilon e_j - \varepsilon e_i,
\]
where \(e_i\) and \(e_j\) denote the standard basis vectors. Then \(p(\varepsilon) \in \Delta^K\). The change in objective value is
\[
\Phi_t(p(\varepsilon)) - \Phi_t(p^\star)
=
\varepsilon \big(s_j^{(t)} - s_i^{(t)}\big)
+
T \Big(
\mathcal{H}(p(\varepsilon)) - \mathcal{H}(p^\star)
\Big).
\]
Only coordinates \(i\) and \(j\) change, so the entropy difference is
\[
\mathcal{H}(p(\varepsilon)) - \mathcal{H}(p^\star)
=
-\varepsilon \log \varepsilon
-
(p_i^\star - \varepsilon)
\log(p_i^\star - \varepsilon)
+
p_i^\star \log p_i^\star.
\]
Using a first-order expansion around \(p_i^\star > 0\), we have
\[
-
(p_i^\star - \varepsilon)
\log(p_i^\star - \varepsilon)
+
p_i^\star \log p_i^\star
=
\varepsilon(\log p_i^\star + 1) + O(\varepsilon^2).
\]
Therefore,
\[
\mathcal{H}(p(\varepsilon)) - \mathcal{H}(p^\star)
=
-\varepsilon \log \varepsilon
+
\varepsilon(\log p_i^\star + 1)
+
O(\varepsilon^2).
\]
Substituting this into the objective difference gives
\[
\Phi_t(p(\varepsilon)) - \Phi_t(p^\star)
=
-T\varepsilon \log \varepsilon
+
\varepsilon
\Big[
s_j^{(t)} - s_i^{(t)}
+
T(\log p_i^\star + 1)
\Big]
+
O(\varepsilon^2).
\]
As \(\varepsilon \to 0^+\), the term \(-\varepsilon \log \varepsilon\) is positive and dominates all linear \(O(\varepsilon)\) terms. Since \(T > 0\), it follows that for sufficiently small \(\varepsilon > 0\),
\[
\Phi_t(p(\varepsilon)) > \Phi_t(p^\star),
\]
which contradicts the optimality of \(p^\star\). Hence, no coordinate of \(p^\star\) can be zero, and therefore \(p_k^\star > 0\) for all \(k\).
\end{proof}

\begin{proposition}[First-order optimality conditions]
\label{prop:entropy-foc}
Let \(p^\star\) be a maximizer of Equation \ref{eq:entropy-priority-allocation}. Then there exists a scalar \(\nu \in \mathbb{R}\) such that, for every \(k = 1, \ldots, K\),
\[
s_k^{(t)}
-
T(\log p_k^\star + 1)
+
\nu
=
0.
\label{eq:entropy-foc}
\]
\end{proposition}

\begin{proof}
By Lemma~Equation \ref{lem:entropy-interior}, any maximizer \(p^\star\) lies in the interior of the simplex. Hence, the non-negativity constraints are inactive at \(p^\star\), and we only need to enforce the equality constraint \(\sum_{k=1}^K p_k = 1\).

The Lagrangian is
\[
\mathcal{L}(p,\nu)
=
\sum_{k=1}^K p_k s_k^{(t)}
-
T \sum_{k=1}^K p_k \log p_k
+
\nu
\left(
\sum_{k=1}^K p_k - 1
\right).
\]
Stationarity at \(p^\star\) requires
\[
\frac{\partial \mathcal{L}}{\partial p_k}(p^\star,\nu)
=
0,
\qquad
k = 1, \ldots, K.
\]
Since
\[
\frac{\partial}{\partial p_k}
\left(
-p_k \log p_k
\right)
=
-(\log p_k + 1),
\]
we obtain
\[
s_k^{(t)}
-
T(\log p_k^\star + 1)
+
\nu
=
0,
\]
which proves the claim.
\end{proof}

\begin{theorem}[Closed-form solution of entropy-regularized priority allocation]
\label{thm:entropy-softmax}
For any \(T > 0\), the optimization problem in Equation \ref{eq:entropy-priority-allocation} has a unique maximizer \(p^{(t),\star} \in \Delta^K\), given by
\[
p_k^{(t),\star}
=
\frac{
\exp\left(s_k^{(t)}/T\right)
}{
\sum_{j=1}^K \exp\left(s_j^{(t)}/T\right)
},
\qquad
k = 1, \ldots, K.
\label{eq:entropy-softmax}
\]
\end{theorem}

\begin{proof}
Existence follows from Lemma~\ref{lem:entropy-existence}. By Lemma~\ref{lem:entropy-interior} and Proposition~\ref{prop:entropy-foc}, any maximizer \(p^\star\) satisfies
\[
s_k^{(t)}
-
T(\log p_k^\star + 1)
+
\nu
=
0
\]
for some scalar \(\nu \in \mathbb{R}\). Rearranging gives
\[
\log p_k^\star
=
\frac{s_k^{(t)} + \nu}{T}
-
1.
\]
Exponentiating both sides yields
\[
p_k^\star
=
\exp\left(\frac{s_k^{(t)}}{T}\right)
\exp\left(\frac{\nu}{T} - 1\right).
\]
The factor \(\exp(\nu/T - 1)\) is independent of \(k\). Therefore,
\[
p_k^\star
\propto
\exp\left(\frac{s_k^{(t)}}{T}\right).
\]
Enforcing the simplex constraint \(\sum_{k=1}^K p_k^\star = 1\) gives
\[
p_k^\star
=
\frac{
\exp\left(s_k^{(t)}/T\right)
}{
\sum_{j=1}^K \exp\left(s_j^{(t)}/T\right)
},
\]
which proves the closed form.

It remains to show uniqueness. The map
\[
p \mapsto \langle p, s^{(t)} \rangle
\]
is linear, and the entropy map \(p \mapsto \mathcal{H}(p)\) is strictly concave on the interior of \(\Delta^K\). Since \(T > 0\), \(\Phi_t\) is strictly concave on the interior of \(\Delta^K\). By Lemma~\ref{lem:entropy-interior}, every maximizer lies in the interior. Therefore, there cannot be two distinct maximizers, and the maximizer is unique.
\end{proof}

\begin{proposition}[RECAP reweighting rule]
\label{prop:recap-reweighting}
Define the RECAP loss weights by
\[
\lambda_k^{(t)}
:=
K p_k^{(t),\star}.
\]
Then
\[
\lambda_k^{(t)}
=
\frac{
K \exp\left(s_k^{(t)}/T\right)
}{
\sum_{j=1}^K \exp\left(s_j^{(t)}/T\right)
},
\qquad
k = 1, \ldots, K,
\label{eq:recap-reweighting}
\]
and the weights satisfy the mean-normalization property
\[
\frac{1}{K}
\sum_{k=1}^K
\lambda_k^{(t)}
=
1.
\]
\end{proposition}

\begin{proof}
Substituting the closed-form optimizer from Theorem~\ref{thm:entropy-softmax} into the definition \(\lambda_k^{(t)} := K p_k^{(t),\star}\) gives
\[
\lambda_k^{(t)}
=
K
\frac{
\exp\left(s_k^{(t)}/T\right)
}{
\sum_{j=1}^K \exp\left(s_j^{(t)}/T\right)
}
=
\frac{
K \exp\left(s_k^{(t)}/T\right)
}{
\sum_{j=1}^K \exp\left(s_j^{(t)}/T\right)
}.
\]
Moreover,
\[
\frac{1}{K}
\sum_{k=1}^K
\lambda_k^{(t)}
=
\frac{1}{K}
\sum_{k=1}^K
K p_k^{(t),\star}
=
\sum_{k=1}^K
p_k^{(t),\star}
=
1,
\]
where the last equality follows from \(p^{(t),\star} \in \Delta^K\). This proves the proposition.
\end{proof}

\subsection{Data statistics and implementation details}
 Table~\ref{tab:stats} reports the full statistics of the training corpora used in our experiments.
For {LLaVA-OneVision-OCR}, we extract OCR-focused subsets from the official {LLaVA-OneVision} release~\citep{li2024llava}: {IIIT5K}~\citep{mishra2012iiit5k}, {HME100K}~\citep{hme100k}, {IAM}~\citep{marti2002iam}, {TextCaps}~\citep{sidorov2020textcaps}, and {TextOCR}~\citep{singh2021textocr}, alongside release-provided synthetic/curated subsets (rendered\_text, k12\_printing, chrome\_writing). These images are resized so that the longer side is $\leq 512$\,px while preserving aspect ratio to mitigate out-of-memory errors without altering task semantics.

\begin{table}[!ht]
  \centering
  \caption{\label{tab:stats}\textbf{Data statistics for each data source.} We present the original volume of data (\# samples).
  }
  \label{tab:datasets}
  \resizebox{\textwidth}{!}{\begin{tabular}{l|lllcr}
    \toprule
    \textbf{Dataset} &
    \textbf{Domain} &
    \textbf{Answer Type} &
    \textbf{Rewards/Objectives} &
    \textbf{\#\,samples} \\
    \midrule

RefCOCO  \citep{kazemzadeh-etal-2014-referitgame} & Referring Expression Comprehension & 2D Bounding Box & IoU, Answer Format &      321327   \\
LLaVA-OneVision-OCR \citep{li2024llava} &  Scene Text-Centric Visual Question Answering    & Natural Language & Next Token Prediction   & 66468      \\
ThinkLite-VL-70k \citep{wang2025sota}  & Math Reasoning \& Natural Image/Chart Understanding  & Natural Language & Acc, Thinking Format    & 69997      \\
     LISA-train~\citep{lai2023lisa} & Referring Expression  & 2D Bounding Box & IoU, Thinking Format & 1326 \\
     GeoQAV~\citep{li2025openr1multimodal} & Math Visual Question Answering &  Multiple Choice & Acc, Thinking Format & 1969 \\
     SAT-train~\citep{ray2024satspatialaptitudetraining} & Spatial Visual Question Answering & Natural Language & Acc, Thinking Format & 15000 \\
      ScienceQA-train~\citep{scienceqa} & Science Visual Question Answering & Multiple Choice & Acc, Thinking Format & 6218 \\

    \bottomrule
  \end{tabular}
  }
\end{table}

We optimize with GRPO and SFT losses using AdamW \citep{loshchilov2017decoupled} ($\beta_1{=}0.9$, $\beta_2{=}0.999$, $\varepsilon{=}10^{-8}$). The learning rate follows a linear schedule: $10\%$ warm-up to $\eta_{\max}{=}1{\times}10^{-6}$, then linear decay to $0$. Window size $W$ and temperature $T$ are set to 10 and 5.0, respectively, in our experiments. Due to the large computational requirements of RL training, we find that setting $T = 5$ and $\alpha = 0.5$ works reasonably well in the RLVR-only setting. For simplicity, we keep this configuration for the hybrid setup and do not perform additional hyperparameter tuning in the large-scale setting. All runs use {bfloat16} precision \citep{wang2019bfloat16, micikevicius2018mixed} and FlashAttention kernels \citep{dao2022flashattention} for memory- and throughput-efficient attention. We enable thinking mode on reasoning tasks by enforcing structured traces (i.e., wrapping thoughts in \texttt{<think>}...\texttt{</think>}), which has been shown to improve reasoning and transparency~\citep{hu2025beyond,xie2025logic,chen2025reasoning}.

For the larger hybrid setting, each model is trained for 500 steps on 8 GPUs using data parallelism, with a per-device batch size of 1, 2 gradient accumulation steps (effective batch size 16), and 4 rollouts per prompt (64 rollouts per optimizer step). We evaluate our models on a broad suite of widely used VLM benchmarks spanning general multimodal understanding, visual reasoning, math-in-vision, OCR, and accessibility: {LISA} \citep{Lai2024LISA}, {MMMU-Pro} \citep{Yue2024MMMUPro}, {AI2D} \citep{Kembhavi2016AI2D}, {MathVista} \citep{Lu2023MathVista}, {MathVision} \citep{Liu2024MATHVision}, {MathVerse} \citep{Zhang2024MathVerse}, {MMBench} \citep{Liu2024MMBench}, {VizWiz} \citep{Gurari2018VizWiz}, and {OCRBench v2} \citep{Fu2024OCRBenchV2}. For the smaller setup, we also follow \citet{liang2025modomodomultidomaindatamixtures} and train on 8 GPUs using data parallelism, with a per-device batch size of 2 and 4 rollouts per prompt, and then evaluate the models on {SAT} \citep{Ray2024SAT}, {ScienceQA} \citep{Lu2022ScienceQA}, {MathVista} \citep{Lu2023MathVista}, {ChartQA} \citep{Masry2022ChartQA}, {InfoVQA} \citep{Mathew2022InfographicVQA}, and {MMMU} \citep{Yue2023MMMU}. Our evaluation protocol closely follows LMMS-Eval \citep{zhang2024lmmsevalrealitycheckevaluation} and VLMEvalKit \citep{duan2024vlmevalkit}. 

\label{eval-prompt}
\begin{tcolorbox}[
    colback=white!95!blue,
    colframe=white!30!blue,
    title=Evaluation prompt,
    fonttitle=\bfseries\large
]
\textbf{Non-Thinking:} \\
\{Question\} \\ Output the answer in \texttt{<answer>} \texttt{</answer>} tags.\\
\textbf{Thinking:} \\
\{Question\} \\ Output the thinking process in \texttt{<think>} \texttt{</think>} tags and the final answer (option) in \texttt{<answer>} \texttt{</answer>} tags.
\end{tcolorbox}

\begin{algorithm}[t]
\caption{Replay-Enhanced CApability Preservation (RECAP).}
\label{alg:recap}
\begin{algorithmic}[1]
\REQUIRE Base parameters $\theta^{(0)}$; domain list $\{\mathcal{D}_n\}_{n=1}^{N}$ comprising general domains $\{\mathcal{D}^G_1,\dots,\mathcal{D}^G_M\}$ and reasoning domains $\{\mathcal{D}^R_1,\dots,\mathcal{D}^R_L\}$; objectives $\{L_k\}_{k=1}^K$; window size $W$; temperature $T$; total iterations $T_{\max}$
\STATE Initialize $\lambda_k^{(0)} \gets 1$ for all $k$
\STATE Initialize loss history buffers $\mathcal{B}_k$ of length $2W$ for each objective $k$
\FOR{$t = 1$ to $T_{\max}$}
    \STATE Sample mini-batches from reasoning and replay data on each domain $\mathcal{D}_n$
    \STATE Compute per-domain, per-objective losses $\ell_{n,k}^{(t)}(\theta^{(t)})$
    \STATE Compute per-objective averaged losses
    \[
        L_k^{(t)} \gets \frac{1}{N}\sum_{n=1}^{N} \ell_{n,k}^{(t)}(\theta^{(t)}), \quad \forall k
    \]
    \FOR{$k = 1$ to $K$}
        \STATE Push $L_k^{(t)}$ into buffer $\mathcal{B}_k$ (FIFO)
    \ENDFOR
    \IF{$t \ge 2W$}
        \FOR{$k = 1$ to $K$}
            \STATE Compute current-window mean:
            $
                \mu_k^{(t)} \gets \frac{1}{W} \sum_{s=t-W+1}^{t} L_k^{(s)}
            $
            and previous-window mean $
                \tilde{\mu}_k^{(t)} \gets \frac{1}{W} \sum_{s=t-2W+1}^{t-W} L_k^{(s)}
            $
            \STATE Compute the instability:
            \[
                \sigma_k^{(t)} =
      \sqrt{\frac{1}{2W-1}\sum_{s=t-2W+1}^{t}\bigl(L_k^{(s)}-\mu_k^{(t)}\bigr)^2}
            \]
            \STATE Compute the convergence rate
            $
                c_k^{(t)} \gets \frac{\tilde{\mu}_k^{(t)}}{{\mu}_k^{(t)} }
            $
            , the inverse signal-to-noise ratio
            $
                i_k^{(t)} \gets
                \frac{\sigma_k^{(t)}}{\mu_k^{(t)} + \tilde{\mu}_k^{(t)}}
            $
            and the relative priority between domains:
            \[
                s_k^{(t)} \gets c_k^{(t)} + i_k^{(t)}
            \]
        \ENDFOR
        \STATE Calculate softmax weights:
        \[
            \lambda_k^{(t)} \gets
            \frac{K\,\exp\!\bigl(s_k^{(t)}/T\bigr)}
                 {\sum_{j=1}^{K}\exp\!\bigl(s_j^{(t)}/T\bigr)},
            \quad \forall k
        \]
    \ELSE
        \STATE $\lambda_k^{(t)} \gets 1$ for all $k$
    \ENDIF
    \STATE Compute final objective:
    \[
        \mathcal{L}^{(t)}(\theta^{(t)}) \gets \frac{1}{K}\sum_{k=1}^{K}\lambda_k^{(t)} L_k^{(t)}
    \]
    \STATE Update parameters:
    \[
        \theta^{(t+1)} \gets \theta^{(t)} - \eta \nabla_{\theta} \mathcal{L}^{(t)}(\theta^{(t)})
    \]
\ENDFOR
\end{algorithmic}
\end{algorithm}

We provide the description of RECAP in Algorithm \ref{alg:recap} and its pseudocode in Algorithm~\ref{alg:pseudo}. To use it, one first computes the task losses, calls {\PyCode{update}} to update the task weighting, and then obtains the weighted loss via {\PyCode{get\_weighted\_loss}} to perform standard backpropagation. For typical settings (e.g., Qwen-3B/7B, $K<10$ objectives, window size $W=10$), our method introduces only $\Theta(K W)$ extra scalar operations and $\Theta(K W)$ memory, which is negligible compared to the $\Theta\left(10^{11}\right)-\Theta\left(10^{12}\right)$ FLOPs per step of the underlying model; in practice, we observed no measurable slowdown.

\begin{algorithm}[t!]
\begin{flushleft}
\PyCode{class RECAP:}\\
\PyCode{def \_\_init\_\_(self, num\_objectives, window\_size, T=5.0):}\\
\qquad\PyComment{num\_objectives~~(K,) number of objectives (rewards / losses)}\\
\qquad\PyComment{window\_size~~W, length of each averaging window}\\
\PyCode{~~~~self.K = num\_objectives}\\
\PyCode{~~~~self.W = window\_size}\\
\PyCode{~~~~self.T = T}\\
\PyComment{~~~~loss\_history[k] stores a list of recent scalar losses for objective k}\\
\PyCode{~~~~self.loss\_history = [deque(maxlen=2 * self.W)}\\
\PyCode{~~~~~~~~~~~~~~~~~~~~~~~~~~~~for \_ in range(num\_objectives)]}\\
\PyComment{~~~~current weights \(\lambda\) (no grad, treated as buffer)}\\
\PyCode{~~~~self.lambdas = torch.ones(self.K)}\\
\PyCode{~~~~self.step = 0}\\
\PyCode{}\\
\PyCode{def get\_weighted\_loss(self, losses):}\\
\qquad\PyComment{losses~~(K,) tensor of per-objective losses \(L_k^{(t)}\)}\\
\qquad\PyComment{returns \(\mathcal{L}^{(t)} = \frac{1}{K}\sum_k \lambda_k^{(t)} L_k^{(t)}\)}\\
\PyCode{~~~~weights = self.lambdas.detach()}\\
\PyCode{~~~~loss = (weights * losses).mean()}\\
\PyCode{~~~~return loss}\\
\PyCode{}\\
\PyCode{def update(self, losses):}\\
\qquad\PyComment{losses~~(K,) tensor of current per-objective losses (no grad needed)}\\
\PyCode{~~~~self.step += 1}\\
\PyComment{~~~~append current losses into history (FIFO of length at most 2W)}\\
\PyCode{~~~~for k in range(self.K):}\\
\PyCode{~~~~~~~~self.loss\_history[k].append(losses[k].detach())}\\
\PyCode{}\\
\PyComment{~~~~if not enough history, keep uniform mixing}\\
\PyCode{~~~~if self.step < 2 * self.W:}\\
\PyCode{~~~~~~~~self.lambdas = torch.ones(self.K)}\\
\PyCode{~~~~~~~~return}\\
\PyCode{}\\
\PyComment{~~~~compute per-objective signals \(c_k^{(t)}\) and \(i_k^{(t)}\)}\\
\PyCode{~~~~c = torch.zeros(self.K)}\\
\PyCode{~~~~i = torch.zeros(self.K)}\\
\PyCode{~~~~for k in range(self.K):}\\
\PyCode{~~~~~~~~hist = torch.stack(self.loss\_history[k])}  \\
\PyCode{~~~~~~~~recent = hist[-self.W:]}  \PyComment{current window} \\
\PyCode{~~~~~~~~old = hist[-2*self.W:-self.W]} \PyComment{previous window}\\
\PyCode{~~~~~~~~mu = recent.mean()}\\
\PyCode{~~~~~~~~mu\_old = old.mean()}\\
\PyCode{~~~~~~~~sigma = hist.std(unbiased=True)}\\
\PyComment{~~~~~~~~\(c_k^{(t)} =   \tilde{\mu}_k^{(t)} / \mu_k^{(t)}\)}\\
\PyCode{~~~~~~~~c[k] = mu\_old / mu }\\
\PyComment{~~~~~~~~\(i_k^{(t)} = \sigma_k^{(t)} / (\mu_k^{(t)} + \tilde{\mu}_k^{(t)})\)}\\
\PyCode{~~~~~~~~i[k] = sigma / (mu + mu\_old)}\\
\PyCode{}\\
\PyComment{~~~~priority scores \(s_k^{(t)} = c_k^{(t)} + i_k^{(t)}\)}\\
\PyCode{~~~~s = c + i}\\
\PyComment{~~~~temperature-controlled softmax, normalized so \(\frac{1}{K}\sum_k \lambda_k = 1\)}\\
\PyCode{~~~~w = torch.softmax(s / self.T, dim=-1)}\\
\PyCode{~~~~self.lambdas = self.K * w.detach()}\\
\end{flushleft}
\caption{Implementation of our proposed method in PyTorch-like pseudocode}
\label{alg:pseudo}
\end{algorithm}

\clearpage
\subsection{Prompts used in our experiments}
\begin{figure}[!ht]
    \centering
     \includegraphics[width=.8\textwidth]{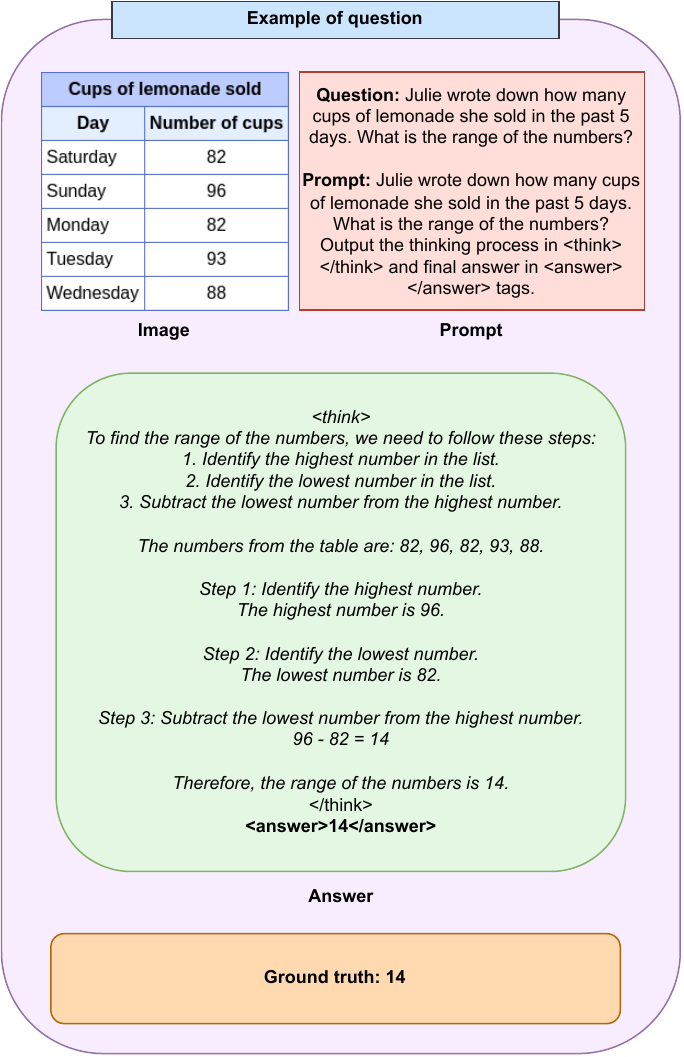}
    \caption{\textbf{Example conversation for finetuning.} We ask the model to produce its reasoning chain within \texttt{<think>} $\dots$ \texttt{</think>} and the final answer in \texttt{<answer>}$\dots$\texttt{</answer>} tags. }
    \label{fig:example_qa}
\end{figure}

The prompt used for training our model is shown in Figure \ref{fig:example_qa}.

\clearpage
\subsection{Qualitative examples}
\begin{figure}[!ht]
    \centering
     \includegraphics[width=1.\textwidth]{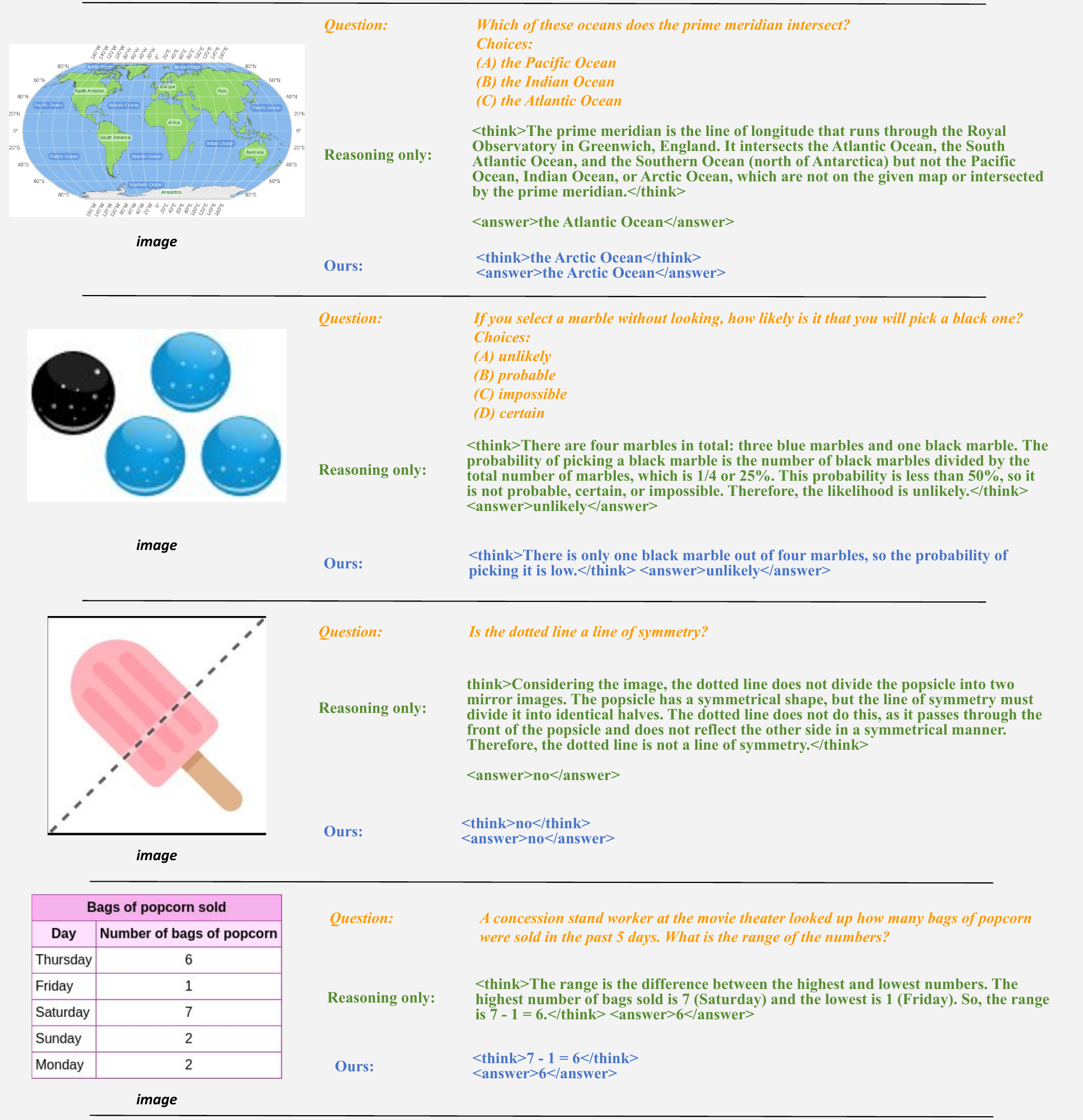}
    \caption{\textbf{Generated answers from the reasoning-only finetuned model and our model.} While the reasoning-only finetuned model generates long reasoning chains for simple visual questions, our model produces more concise answers, especially on non-mathematical tasks.}
    \label{fig:short-response}
\end{figure}

Figure \ref{fig:short-response} presents responses from models trained either with reasoning data only or with additional general data.

\clearpage
\subsection{Evolution of rollouts during training with the thinking reward}

Below, we show generated responses from the model at different points during training, where we observe the finetuned model gradually shortening its chain of thought on image-understanding tasks to provide final answers directly. This motivates our use of non-thinking rewards on tasks that do not require reasoning, instead of employing the same thinking-format reward for every task \citep{liang2025modomodomultidomaindatamixtures}.

\newtcbtheorem{Example}{\bfseries Example}{enhanced,drop shadow={black!50!white},
  coltitle=black,
  top=0.3in,
  attach boxed title to top left=
  {xshift=1.5em,yshift=-\tcboxedtitleheight/2},
  boxed title style={size=small,colback=pink}
}{Example}

\begin{Example}{rollout \# 0}{}
\textbf{Prompt:} Please provide the bounding box coordinate of the region this sentence describes: flicking tail.

Output the thinking process in \texttt{<think>} \texttt{</think>} and final answer in \texttt{<answer>} \texttt{</answer>} tags. The output answer format should be as follows:

\texttt{<answer>}[x1, y1, x2, y2]\texttt{</answer>}

\textbf{Answer:} To determine the bounding box coordinates for the zebra that is flicking its tail, we need to follow these steps:

1. Identify the zebra in the image that is flicking its tail.

2. Observe the position and direction of the zebra's tail.

3. Define a bounding box that encapsulates the entire zebra, keeping in mind that the tail should be included in the box.

Upon inspection:

- The zebra on the right is flicking its tail, as evident from the blurred tail motion and the position of the tail.

- The bounding box should encompass the entire zebra, including its tail.

\texttt{<answer>}[291, 41, 432, 315]\texttt{</answer>}
\end{Example}

\begin{Example}{rollout \# 500}{Example 2}
\textbf{Prompt:} Please provide the bounding box coordinate of the region this sentence describes: second from right.

Output the thinking process in \texttt{<think>} \texttt{</think>} and final answer in \texttt{<answer>} \texttt{</answer>} tags. The output answer format should be as follows:

\texttt{<answer>}[x1, y1, x2, y2]\texttt{</answer>}

\textbf{Answer:} \texttt{<think>}To determine the bounding box for the second from right, I first identify the rightmost object, which is the one on the far right, identified by its position and bounding box. The second from right is the one to the left of the rightmost object. By visual inspection, the bounding box for the second from right should start after the farthest object and extend leftward.\texttt{</think>}

\texttt{<answer>}[237, 72, 374, 328]\texttt{</answer>}
\end{Example}

\begin{Example}{rollout \# 1000}{Example 3}
\textbf{Prompt:} Please provide the bounding box coordinate of the region this sentence describes: boat blocked by cow stomach.

Output the thinking process in \texttt{<think>} \texttt{</think>} and final answer in \texttt{<answer>} \texttt{</answer>} tags. The output answer format should be as follows:

\texttt{<answer>}[x1, y1, x2, y2]\texttt{</answer>}

\textbf{Answer:} \texttt{<think>} The sentence refers to the boat that is partially covered by the cow's stomach. This boat seems to be the one with the green and white colors.\texttt{</think>}

\texttt{<answer>}[327,21,631,280]\texttt{</answer>}
\end{Example}

\begin{Example}{rollout \# 5000}{Example 4}
\textbf{Prompt:} Please provide the bounding box coordinate of the region this sentence describes: guy on left with backpack on.

Output the thinking process in \texttt{<think>} \texttt{</think>} and final answer in \texttt{<answer>} \texttt{</answer>} tags. The output answer format should be as follows:

\texttt{<answer>}[x1, y1, x2, y2]\texttt{</answer>}

\textbf{Answer:} \texttt{<think>}[114,53,236,378]\texttt{</think>}

\texttt{<answer>}[114,53,236,378]\texttt{</answer>}
\end{Example}

\subsection{Reward dynamics during training}

\begin{figure}[!ht]
    \centering
     \includegraphics[width=.95\textwidth]{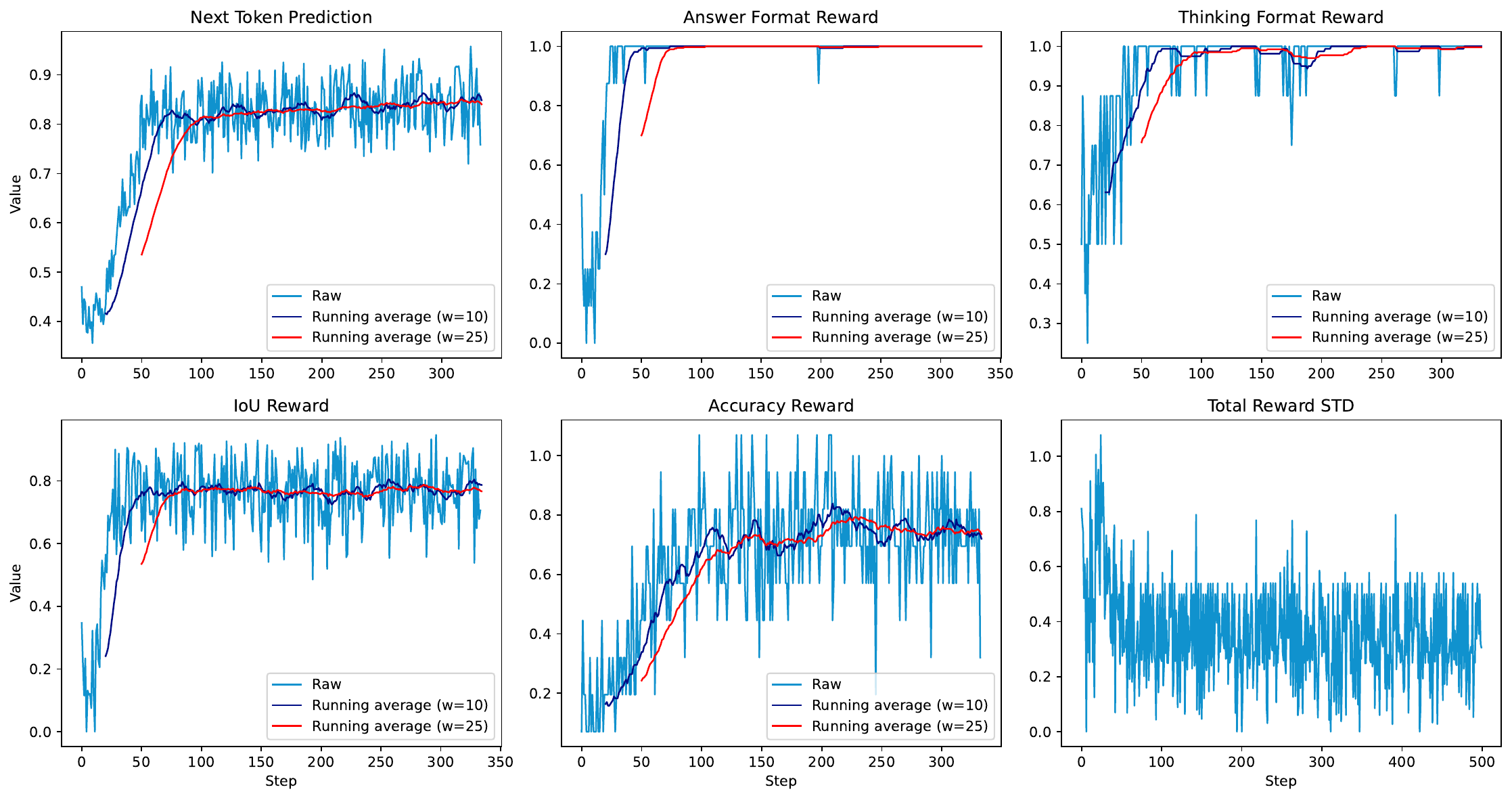}
    \caption{\textbf{Reward dynamics and variability during RLVR training.}
    Per-step rewards (light traces) and sliding-window means (dark curves) for six metrics: {Next-Token Prediction}, {Answer-Format}, {Thinking-Format}, {IoU}, {Accuracy}, and the {Total-Reward Standard Deviation} (lower-right). Asynchronous convergence and high variance motivate short-horizon statistics for dynamic objective reweighting rather than per-iteration magnitudes.{\label{fig:reward_std}}}
\end{figure}

Given the multi-objective nature of the problem, one might apply existing methods in the multi-task learning literature \citep{guo2025revisit} to reweight different objectives and rewards. In practice, this is difficult for two reasons. {First}, computing per-objective gradients is prohibitively expensive at LLM scale, especially under reinforcement learning. {Second}, on-policy RL signals are high-variance and non-stationary \citep{henderson2018deeprlmatters}, making per-iteration statistics unreliable indicators of task progress. As shown in Figure~\ref{fig:reward_std}, all rewards fluctuate substantially within their $[0,1]$ range, with the standard deviation of the total reward peaking near $0.9$ around step $\sim 20$. Thus, we propose a method that uses a sliding window, which provides a more robust proxy for understanding convergence behavior.

We also conduct an ablation on the effect of the window size $W$ by increasing it from 10 (our default throughout the experiments) to larger values, up to 25, as shown in Figure~\ref{fig:reward_std}. Since we train Qwen2.5-VL-7B for 500 iterations, setting $W=25$ delays the onset of dynamic reweighting by 50 iterations, according to Algorithms~\ref{alg:recap} and~\ref{alg:pseudo}, because our method requires $2W$ steps of history. By the time reweighting becomes active, some rewards have already entered a near-converged regime, which reduces the usefulness of the convergence-rate term and makes the scheme rely mostly on the instability term. We therefore choose $W=10$ as a reasonable compromise between sensitivity and robustness: it accumulates enough information while remaining responsive to the current state of training.

\subsection{Reward values at the end of training}

\begin{figure}[!ht]
    \centering
     \includegraphics[width=.9\textwidth]{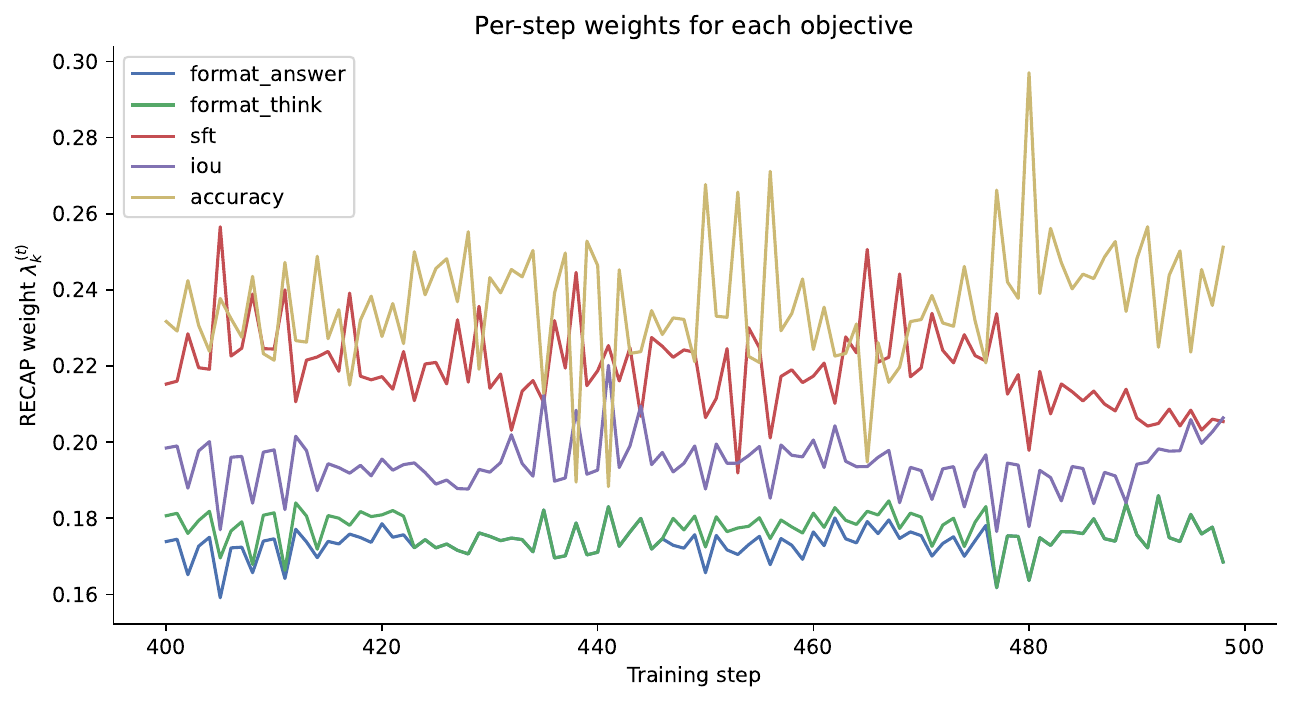}
    \caption{
    \textbf{Evolution of per-objective coefficients.} In the last 100 iterations, the coefficient for each objective is relatively consistent, with format rewards receiving the lowest focus while the supervised finetuning objective and accuracy rewards are emphasized due to their instability.
}
    \label{fig:weight}
\end{figure}

In Figure \ref{fig:weight}, we plot the coefficients of the five objectives used in the hybrid setup. From these coefficients, we can rank the objectives by how strongly our method focuses on each one, from low to high: format rewards, the IoU reward, next-token prediction (on the OCR task), and reasoning accuracy.

\begin{figure}[!ht]
    \centering
     \includegraphics[width=.9\textwidth]{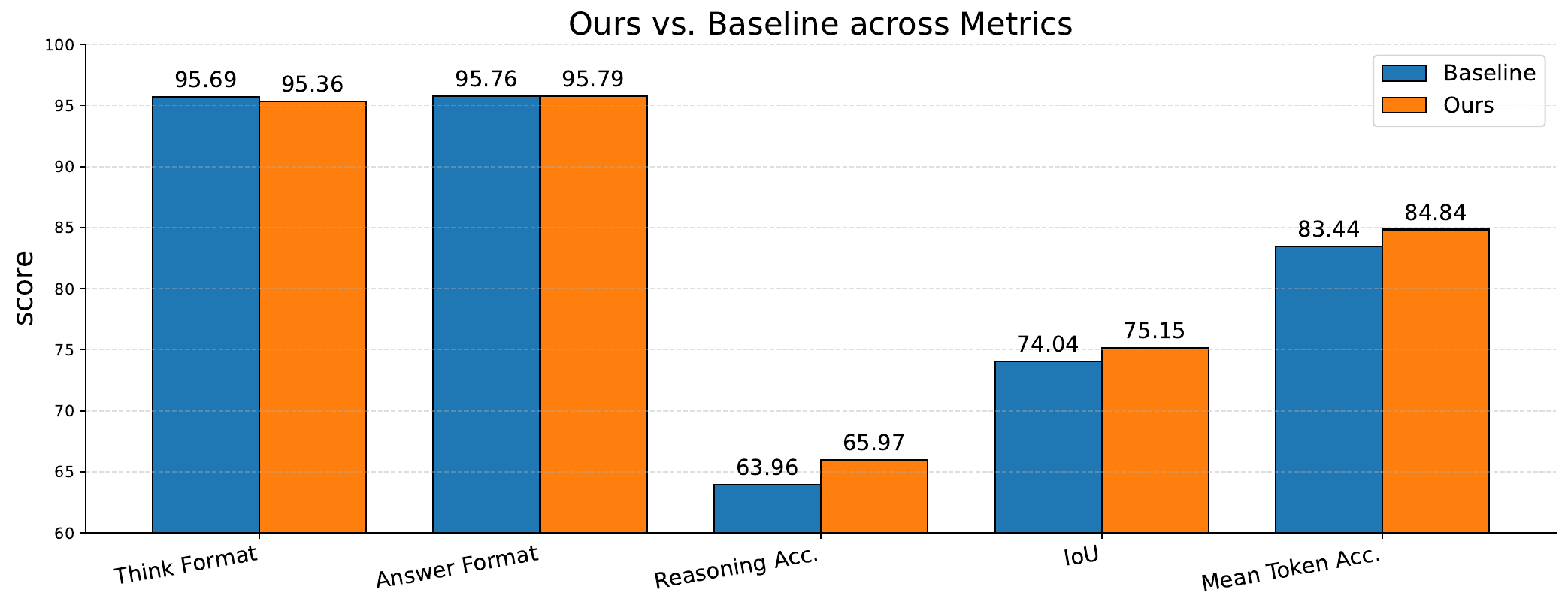}
    \caption{\textbf{Final performance across metrics.} We compare a {uniform} baseline with our {dynamic reweighting}. The gains on correctness-oriented metrics indicate that reallocating weight away from saturated format rewards toward harder objectives yields better solutions without sacrificing adherence to templates.\label{fig:bar}}
\end{figure}

Similar to what we observe in the main paper, Figure \ref{fig:bar} shows the final performance of our model and the uniform baseline. Results show near-parity on thinking formatting $(\texttt{<think>} \texttt{</think>} \;\;  \texttt{<answer>}\;  \texttt{</answer>})$ and direct answer reward ($\texttt{<answer>}\; \texttt{</answer>})$ but consistent improvements in reasoning score, IoU, and mean token accuracy (+2.01, +1.11, and +1.40 points, respectively). This aligns with our design goal: once format signals saturate, we down-weight them and shift capacity to harder, higher-variance objectives, improving accuracy while maintaining output format.

\clearpage

\begin{figure}[!ht]
    \centering
    \begin{subfigure}{0.245\textwidth}
        \centering
        \includegraphics[width=\linewidth,trim=0pt 0pt 0pt 45pt,clip]{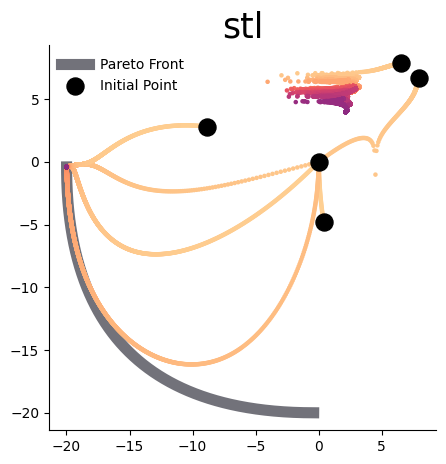}
        \caption{First objective}
        \label{fig:1a}
    \end{subfigure}
    \begin{subfigure}{0.245\textwidth}
        \centering
        \includegraphics[width=\linewidth,trim=0pt 0pt 0pt 45pt,clip]{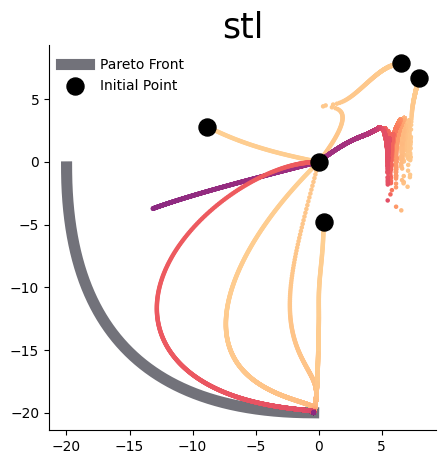}
        \caption{Second objective}
        \label{fig:1b}
    \end{subfigure}
    \begin{subfigure}{0.245\textwidth}
        \centering
        \includegraphics[width=\linewidth,trim=0pt 0pt 0pt 45pt,clip]{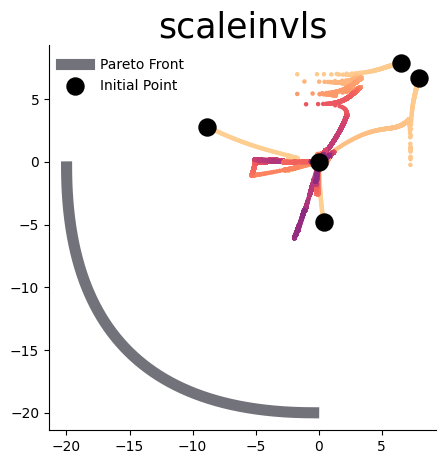}
        \caption{Scale-invariant}
        \label{fig:1c}
    \end{subfigure}
    \begin{subfigure}{0.245\textwidth}
        \centering
        \includegraphics[width=\linewidth,trim=0pt 0pt 0pt 45pt,clip]{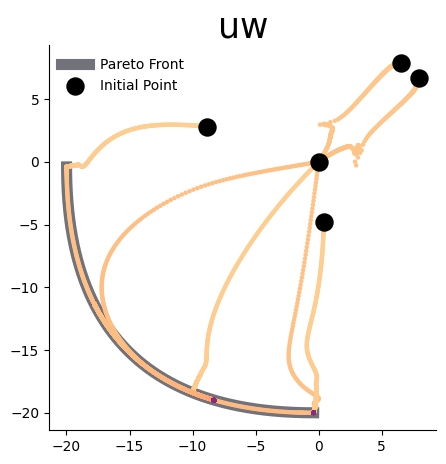}
        \caption{Uncertainty Weighting}
        \label{fig:1d}
    \end{subfigure}
    \begin{subfigure}{0.245\textwidth}
        \centering
        \includegraphics[width=\linewidth,trim=0pt 0pt 0pt 45pt,clip]{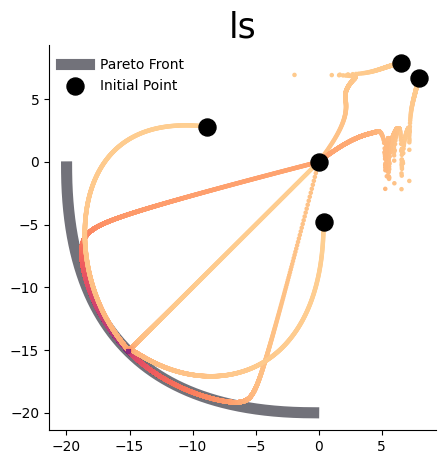}
        \caption{Linear scalarization}
    \end{subfigure}
    \begin{subfigure}{0.245\textwidth}
        \centering
        \includegraphics[width=\linewidth,trim=0pt 0pt 0pt 45pt,clip]{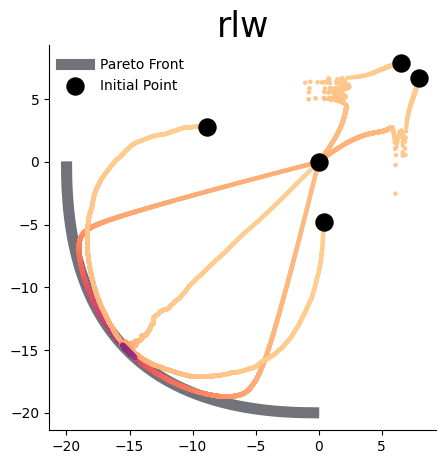}
        \caption{Random loss reweighting}
    \end{subfigure}
    \begin{subfigure}{0.245\textwidth}
        \centering
        \includegraphics[width=\linewidth,trim=0pt 0pt 0pt 45pt,clip]{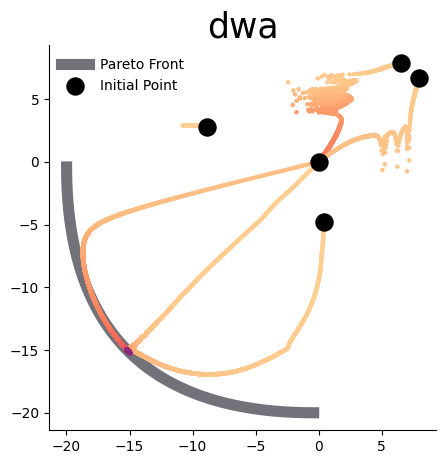}
        \caption{Dynamic Weight Average}
    \end{subfigure}
        \begin{subfigure}{0.245\textwidth}
        \centering
        \includegraphics[width=\linewidth,trim=0pt 0pt 0pt 45pt,clip]{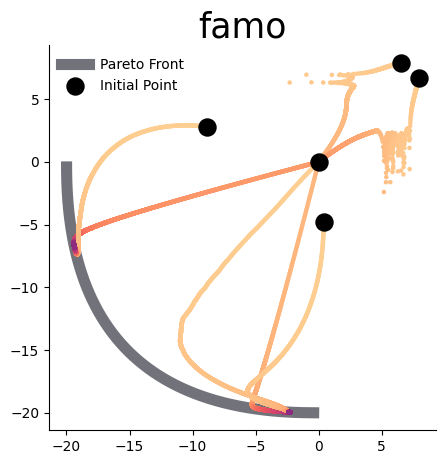}
        \caption{FAMO}
    \end{subfigure}
    \begin{subfigure}{0.245\textwidth}
        \centering
        \includegraphics[width=\linewidth,trim=0pt 0pt 0pt 45pt,clip]{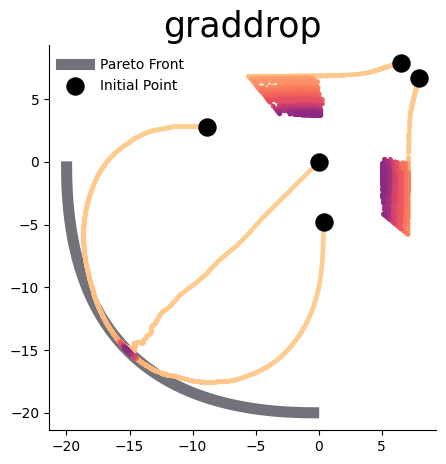}
        \caption{Grad Drop}
    \end{subfigure}
    \begin{subfigure}{0.245\textwidth}
        \centering
        \includegraphics[width=\linewidth,trim=0pt 0pt 0pt 45pt,clip]{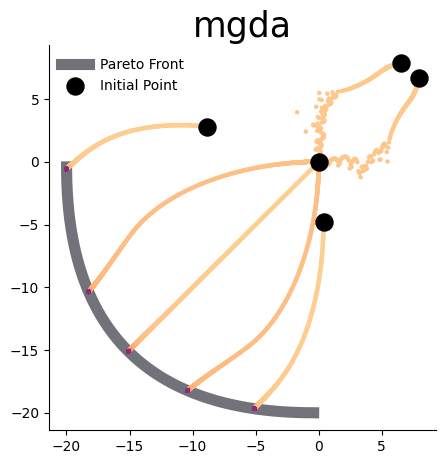}
        \caption{MGDA}
    \end{subfigure}
        \begin{subfigure}{0.245\textwidth}
        \centering
        \includegraphics[width=\linewidth,trim=0pt 0pt 0pt 45pt,clip]{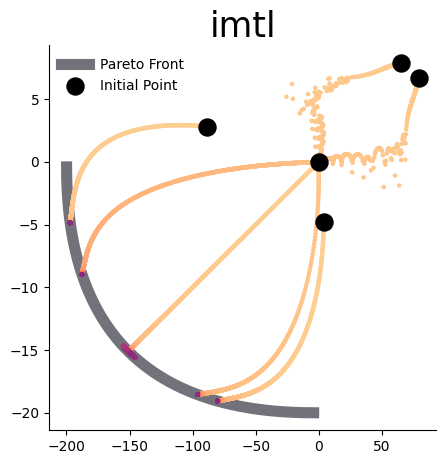}
        \caption{MT-SGD}
    \end{subfigure}
            \begin{subfigure}{0.245\textwidth}
        \centering
        \includegraphics[width=\linewidth,trim=0pt 0pt 0pt 45pt,clip]{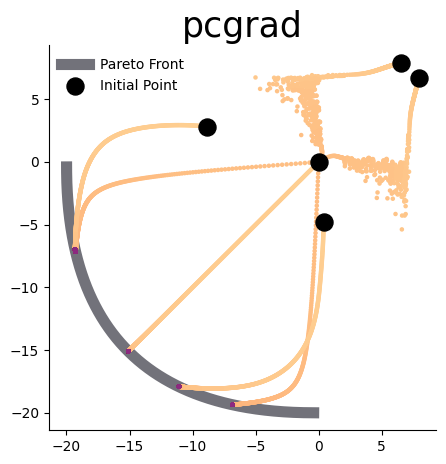}
        \caption{PCGRAD}
        \label{fig:1e}
    \end{subfigure}
    \begin{subfigure}{0.245\textwidth}
        \centering
        \includegraphics[width=\linewidth,trim=0pt 0pt 0pt 45pt,clip]{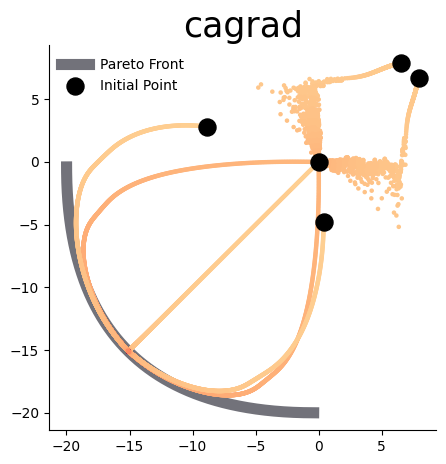}
        \caption{CAGRAD}
        \label{fig:1f}
    \end{subfigure}
    \begin{subfigure}{0.245\textwidth}
        \centering
        \includegraphics[width=\linewidth,trim=0pt 0pt 0pt 45pt,clip]{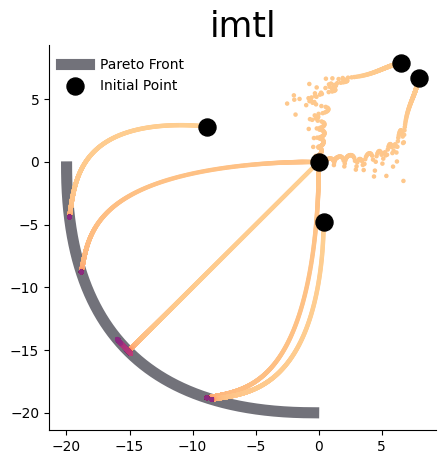}
        \caption{IMTL}
    \end{subfigure}
    \begin{subfigure}{0.245\textwidth}
        \centering
        \includegraphics[width=\linewidth,trim=0pt 0pt 0pt 45pt,clip]{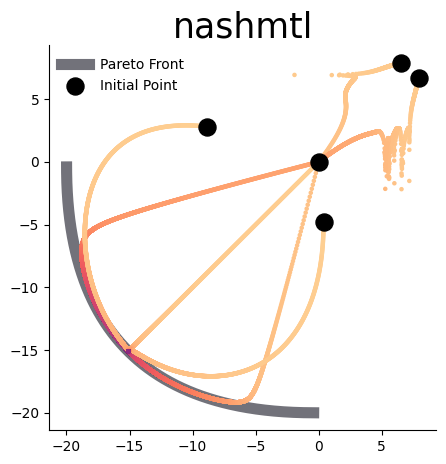}
        \caption{Nash-MTL}
        \label{fig:1g}
    \end{subfigure}
    \begin{subfigure}{0.245\textwidth}
        \centering
        \includegraphics[width=\linewidth,trim=0pt 0pt 0pt 45pt,clip]{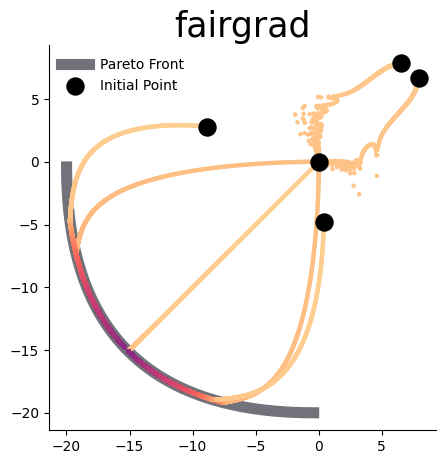}
        \caption{FairGrad}
    \end{subfigure}
    \begin{subfigure}{0.245\textwidth}
        \centering
        \includegraphics[width=\linewidth,trim=0pt 0pt 0pt 45pt,clip]{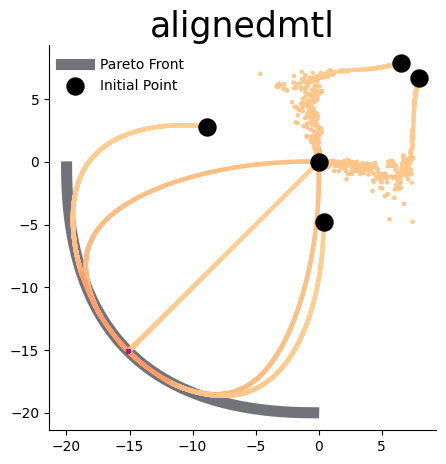}
        \caption{AlignedMTL}
    \end{subfigure}
    \begin{subfigure}{0.245\textwidth}
        \centering
        \includegraphics[width=\linewidth,trim=0pt 0pt 0pt 45pt,clip]{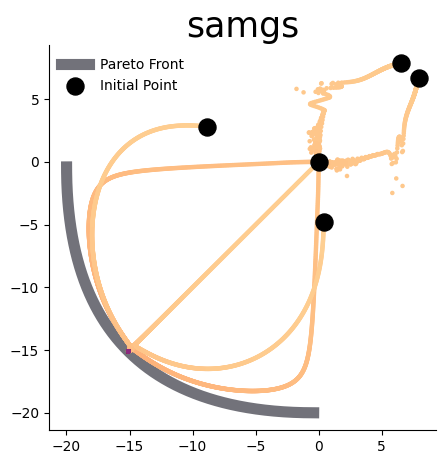}
        \caption{SAMGS}
    \end{subfigure}
    \begin{subfigure}{0.245\textwidth}
        \centering
        \includegraphics[width=\linewidth,trim=0pt 0pt 0pt 45pt,clip]{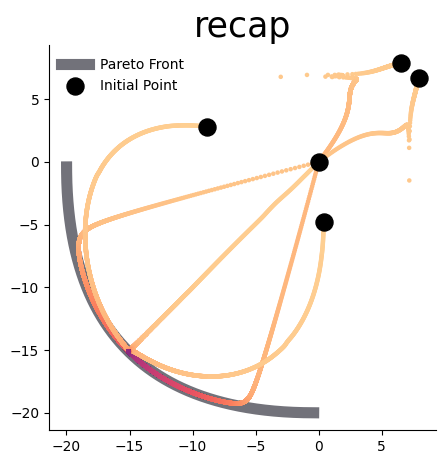}
        \caption{RECAP}
        \label{fig:1h}
    \end{subfigure}
    \caption{A modified illustrative two-task example from \citep{navon2022nashmtl} showing the convergence of comparison methods from different initialization points (black dots $\bullet$). Each optimization trajectory is colored from \textcolor{yellow!60!orange}{yellow} to \textcolor{red}{red}. The bold \textcolor{gray}{gray} line represents the Pareto front. Overall, introducing noise in the first objective causes instability in many MTL methods. Methods that leverage per-step loss-magnitude statistics, such as FAMO, DWA, and UW, exhibit considerable instability during convergence from different initializations. 
    }
    \label{fig:toy-two-task}
\end{figure}

Following the illustrative setup of \citet{navon2022nashmtl}, we consider a synthetic two-task problem with a shared parameter vector and two scalar objectives. The corresponding Pareto front can be computed analytically and is shown in gray in Figure~\ref{fig:toy-two-task}. We benchmark our method against established loss-magnitude-based methods (first row) and gradient-based multi-task learning methods (second row). To mimic the unstable nature of RL training, we inject noise into the first objective, which induces substantial fluctuations for competing methods, whereas our approach remains stable and closely tracks the Pareto front. Additional runtime comparisons in the appendix highlight that our method also achieves favorable wall-clock efficiency, a crucial advantage for large-scale RL training.
The two tasks $L_1(x)$ and $L_2(x)$ are defined on $x=(x_1,x_2)^\top\in\mathbb{R}^2$,
\begin{align*}
    L_1(x) &=  f_1(x)g_1(x)+f_2(x)h_1(x) + 3 \epsilon\\
    L_2(x) &= f_1(x)g_2(x)+f_2(x)h_2(x),
\end{align*}

where $\epsilon \sim N(0,1)$, and the functions are given by:
\begin{align*}
    f_1(x) &= \max\bigl(\tanh(0.5x_2),0\bigr)\\
    f_2(x) &= \max\bigl(\tanh(-0.5x_2),0\bigr)\\
    g_1(x) &= \log\Bigl(\max\bigl(|0.5(-x_1-7)-\tanh(-x_2)|,0.000005\bigr)\Bigr) + 6\\
    g_2(x) &= \log\Bigl(\max\bigl(|0.5(-x_1+3)-\tanh(-x_2)+2|,0.000005\bigr)\Bigr) + 6\\
    h_1(x) &= \bigl((-x_1+7)^2 + 0.1(-x_1-8)^2\bigr)/10 - 20\\
    h_2(x) &= \bigl((-x_1-7)^2 + 0.1(-x_1-8)^2\bigr)/10 - 20.
\end{align*}

 We use five different starting points $\{(-8.5, 7.5), (0, 0), (9.0,9.0), (-7.5,-0.5), (9.0,-1.0)\}$. Those points are optimized with Adam \citep{kingma2014adam} with a learning rate of 1e-2 for 10000 iterations. To justify our proposed reweighting mechanism, we compare against established multi-task learning techniques, including loss-balancing methods such as UW \citep{Kendall2018CVPR_uw}, DWA \citep{Liu2019CVPR_dwa}, GradNorm \citep{Chen2018ICML_gradnorm}, RGW \citep{Lin2022TRML_RGW}, and FAMO \citep{Liu2023NeurIPS_famo}, as well as gradient-based methods: PCGrad \citep{Yu2020NeurIPS_pcgrad}, CAGrad \citep{Liu2021NeurIPS_cagrad}, GradDrop \citep{Chen2020NeurIPS_graddrop}, MGDA \citep{Dong2015_mgda2}, IMTL \citep{Liu2021ICLR_imtl}, MT-SGD \citep{phan2022stochastic}, Nash-MTL \citep{navon2022nashmtl}, FS-MTL \citep{phan2025beyond}, Aligned-MTL \citep{Senushkin2023_alighedmtl}, and SAM-GS \citep{borsani2025gradient}.
Their convergence behavior is presented in Figure \ref{fig:toy-two-task}, where RECAP improves across all initialized solutions compared with other MTL methods. While gradient-based MTL methods such as CAGrad and Nash-MTL do not depend on the initial solutions, they suffer from slow convergence and higher per-step computational cost compared to RECAP.

\clearpage

\begin{table*}[!ht]
\setlength{\tabcolsep}{3pt}
\small
    \centering
    \caption{\textbf{Evaluation results on NYUv2 scene understanding.} Test performance for three tasks: semantic segmentation, depth estimation, and surface normals. We highlight the best loss-magnitude-based MTL method in \textbf{bold} and the best gradient-based MTL method with an \underline{underscore}.}
    \vskip 0.11in
\resizebox{\textwidth}{!}{%
\begin{tabular}{cccccccccccccccccccc}
\toprule\\
 &  &  & \multicolumn{2}{c}{Segmentation} &  & \multicolumn{2}{c}{Depth} &  & \multicolumn{8}{c}{Surface Normal} &  &  &  \\
 \cmidrule(lr){4-5} \cmidrule(lr){7-8} \cmidrule(lr){10-17}
 & Complexity &  & \multirow{2}{*}{mIoU $\uparrow$} & \multirow{2}{*}{Pix Acc $\uparrow$} &  & \multirow{2}{*}{Abs Err $\downarrow$} & \multirow{2}{*}{Rel Err $\downarrow$} &  & \multicolumn{2}{c}{Angle Distance $\downarrow$} &  & \multicolumn{5}{c}{Within $t^\circ$  $\uparrow$} & $\mathbf{\Delta m \%} \downarrow$ &  \\
 \cmidrule(lr){10-11} \cmidrule(lr){13-17}
 &  &  &  &  &  &  &  &  & Mean & Median &  & 11.25 &  & 22.5 &  & 30  &  \\
 \midrule
 & & {STL} & $38.30$ & $63.76$ &  & $0.6754$ & $0.2780$ &  & $25.01$ & $19.21$ &  & $30.14$ &  & $57.20$ &  & $69.15$   &  \\
  \midrule
 & & {LS} & $39.29$ & $65.33$ &  & $0.5493$ & $0.2263$ &  & $28.15$ & $23.96$ &  & $22.09$ &  & $47.50$ &  & $61.08$  & $5.59$ &  \\
 & & {SI} & $38.45$ & $64.27$ && $0.5354$ & $0.2201$ && $27.60$ & $23.37$ && $22.53$ && $48.57$ && $62.32$  & $4.39$  \\
 & & {RLW} & $ 37.17 $ & $ 63.77 $ && $ 0.5759 $ & $ 0.2410 $ && $ 28.27 $ & $ 24.18 $ && $ 22.26 $ && $ 47.05 $ && $ 60.62 $   & $ 7.78 $  \\
 &\(\Theta(1)\) & {DWA} & $39.11$ & $65.31$ &  & $0.5510$ & $0.2285$ &  & $27.61$ & $23.18$ &  & $24.17$ &  & $50.18$ &  & $62.39$& $3.57$ &  \\
 & & {UW} & $36.87$ & $63.17$ &  & $0.5446$ & $0.2260$ &  & $\mathbf{27.04}$ & $22.61$ &  & $23.54$ &  & $49.05$ &  & $63.65$  & $4.05$ &  \\
\rowcolor{carolinablue!60}  & & Ours & \textbf{41.26} &\textbf{66.79}&&\textbf{0.5303}&\textbf{0.2203}&&27.11&\textbf{22.23}&&\textbf{24.64}&&\textbf{50.88}&&\textbf{64.02}& \textbf{0.77} \\
 \midrule
 & & GradNorm &20.09& 64.64&& 0.7200& 0.2800&& $\underline{24.83}$& $\underline{18.86}$&& $\underline{30.8}$ && $\underline{57.94}$&& $\underline{69.73}$ &7.22 \\
 & & {MGDA} & $30.47$ & $59.90$ &  & $0.6070$ & $0.2555$ &  & ${24.88}$ & ${19.45}$ &  & ${29.18}$ &  & ${56.88}$ &  & ${69.36}$  & $1.38$ &  \\
 & & {PCGrad} & $38.06$ & $64.64$ &  & $0.5550$ & $0.2325$ &  & $27.41$ & $22.80$ &  & $23.86$ &  & $49.83$ &  & $63.14$  & $3.97$ &  \\
 & \(\Theta(K)\)& {GradDrop} & $39.39$ & $65.12$ &  & $0.5455$ & $0.2279$ &  & $27.48$ & $22.96$ &  & $23.38$ &  & $49.44$ &  & $62.87$  & $3.58$ &  \\
 & & {CAGrad} & $39.79$ & $65.49$ &  & $0.5486$ & $0.2250$ &  & $26.31$ & $21.58$ &  & $25.61$ &  & $52.36$ &  & $65.58$  & $0.20$ &  \\
 & & {IMTL-G} & $39.35$ & $ 65.60$ &  & $0.5426$ & $0.2256$ &  & $26.02$ & $21.19$ &  & ~~$26.2$ &  & $53.13$ &  & $66.24$  & $ -0.76 $ &  \\
  &  & FS-MTL & $\underline{40.42}$ & $ {65.61}$ &  & ${0.5389}$ & $\underline{0.2121^*}$ &  & ${25.03}$ & ${19.75}$ &  & ~~${28.90}$ &  & ${56.19}$ &  & ${68.72}$ &  $ \underline{-4.77^*} $ &  \\
 & & {Nash-MTL} & ${40.13}$ & $\underline{65.93}$ &  & $\underline{0.5261}$ & ${0.2171}$ &  & $25.26$ & $20.08$ &  & $28.4$ &  & $55.47$ &  & $68.15$  & ${-4.04}$ & \\
 \bottomrule
\end{tabular}%
}
\label{tab:nyu}
\end{table*}

Table~\ref{tab:nyu} reports the performance of different MTL methods on the real-scene understanding benchmark, which includes one segmentation task and two pixel-level regression tasks. Overall, our method nearly matches the single-task baselines ($\mathbf{\Delta m \%} \downarrow \approx 0$) while being roughly $3\times$ more efficient in both runtime and memory, and it consistently outperforms all other loss-reweighting methods across all metrics (except Angle Distance Mean, where it is competitive with Uncertainty Weighting). Notably, our approach even surpasses several established gradient-based methods, such as GradNorm, MGDA, PCGRAD, and GradDrop, while remaining roughly three times faster than gradient-based alternatives. We also observe a clear Pareto trade-off: although NashMTL achieves the highest overall relative improvement in $\mathbf{\Delta m \%} \downarrow$, it lags behind GradNorm and MGDA on the surface-normal task, whereas these methods incur substantial performance drops on segmentation and depth estimation.

\begin{figure}[!ht]
    \centering
     \includegraphics[width=1\textwidth]{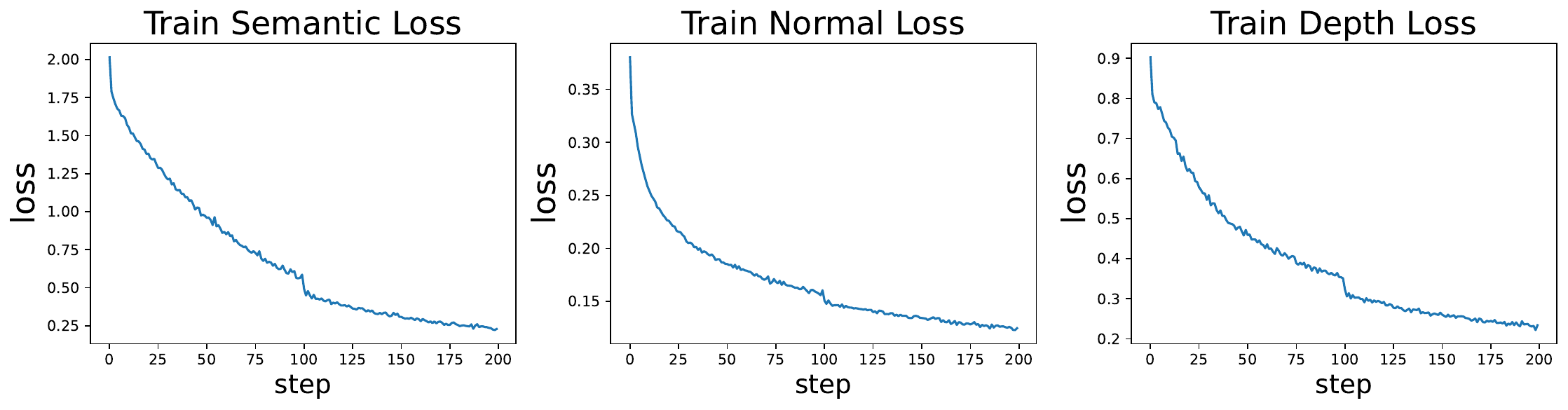}
    \caption{
    \textbf{Loss curves during training on NYUv2.} Compared with the training curves in our experiments (e.g., Figure \ref{fig:reward_std}), these curves are much smoother and more stable, making per-step statistics informative signals of learning progress.
}
    \label{fig:nyu_loss}
\end{figure}

Figure~\ref{fig:nyu_loss} plots the loss curves for three different objectives, showing stable optimization across all of them. In contrast, our RL rewards are much sparser than in this SFT setting, and the training curves in Figures~\ref{fig:reward} and \ref{fig:reward_std} exhibit substantially higher fluctuations. This motivates a more robust loss-reweighting mechanism, as relying solely on instantaneous per-step loss values is not sufficiently representative of the underlying learning dynamics or objective progress.

\begin{figure}[t]
    \centering
     \includegraphics[width=1\textwidth]{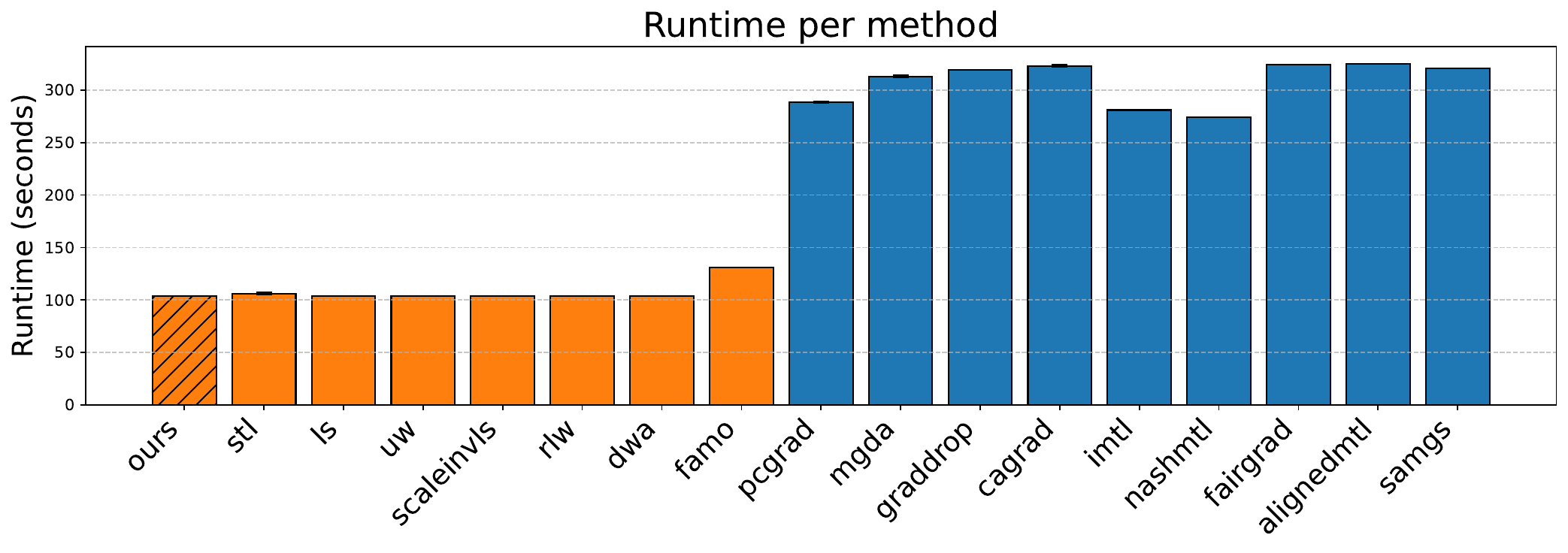}
    \caption{
    \textbf{Running time of different MTL methods.} While robust to noise in some scenarios, gradient-based methods (denoted by \textcolor{blue}{blue}) often incur significant overhead ($\approx K$ times, because they compute per-objective gradients) compared with loss-magnitude-based methods (denoted by \textcolor{orange}{orange}).
}
    \label{fig:time}
\end{figure}

The running-time comparison in Figure~\ref{fig:time} shows that, although effective in some scenarios, gradient-based MTL methods require storing and computing all task gradients, incurring $\Theta(K)$ space and time overhead, where $K$ is the number of objectives. In our illustrative setup with $K=3$, this already makes these methods about three times slower ($\sim 300$s vs. $\sim 100$s) than single-task baselines and other loss-reweighting approaches. In our main RLVR experiments, we have four domains with two objectives per domain ($K=8$), which would make gradient-manipulation methods roughly 8$\times$ slower than standard training. For this reason, we focus on loss-reweighting mechanisms, which avoid such substantial computational overhead.

\begin{figure}[!ht]
    \centering
    \begin{subfigure}{0.26\textwidth}
        \centering
        \includegraphics[width=\linewidth,trim=0pt 0pt 0pt 45pt,clip]{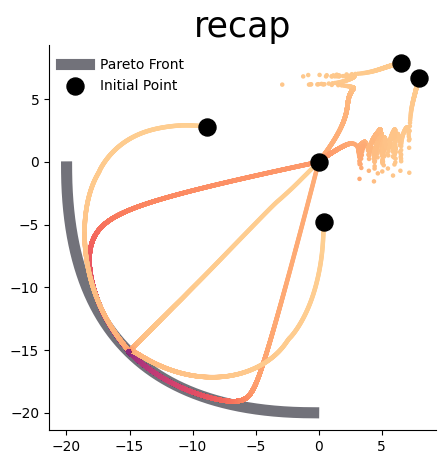}
        \caption{$\alpha$ = 0}
    \end{subfigure}
    \begin{subfigure}{0.26\textwidth}
        \centering
        \includegraphics[width=\linewidth,trim=0pt 0pt 0pt 45pt,clip]{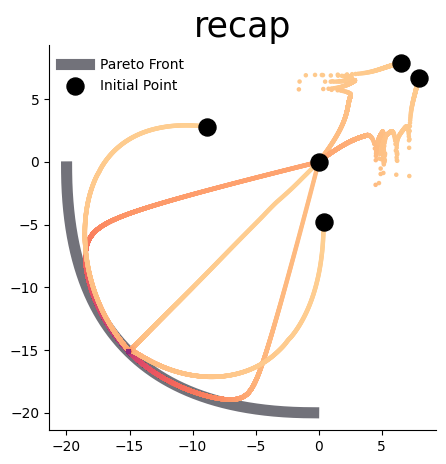}
        \caption{$\alpha$ = 0.25}
    \end{subfigure}
    \begin{subfigure}{0.26\textwidth}
        \centering
        \includegraphics[width=\linewidth,trim=0pt 0pt 0pt 45pt,clip]{figure-new/mtl/recap.png}
        \caption{$\alpha$ = 0.5}
    \end{subfigure}
        \begin{subfigure}{0.26\textwidth}
        \centering
        \includegraphics[width=\linewidth,trim=0pt 0pt 0pt 45pt,clip]{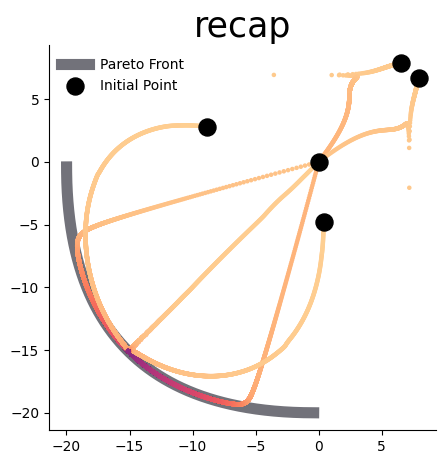}
        \caption{$\alpha$ = 0.75}
    \end{subfigure}
    \begin{subfigure}{0.26\textwidth}
        \centering
        \includegraphics[width=\linewidth,trim=0pt 0pt 0pt 45pt,clip]{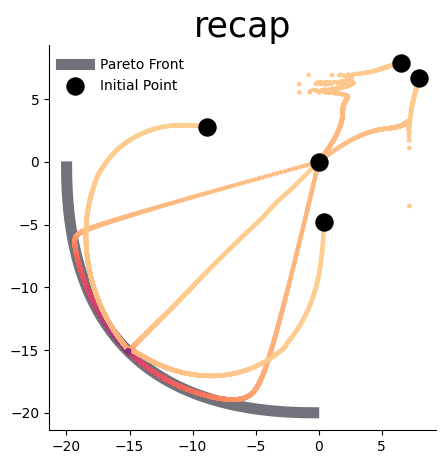}
        \caption{$\alpha$ = 1.}
    \end{subfigure}
    \caption{\textbf{Ablation on the trade-off $\alpha$.} Using only the convergence rate ($\alpha=1$) or only the inverse signal-to-noise ratio ($\alpha=0$) leads to unstable learning for the second and first objectives, respectively.
    \label{fig:tradeoff}}
\end{figure}

We conduct ablation studies on the temperature hyperparameter $T$ and the trade-off $\alpha$ between the convergence rate and the inverse signal-to-noise ratio:  $ s_k^{(t)} = \alpha c_k^{(t)} + (1-\alpha) i_k^{(t)}$. From Figure~\ref{fig:tradeoff}, we observe that intermediate values such as $\alpha=0.5$ or $0.75$ strike a good balance between the two terms and yield noticeably more stable convergence across all initializations.

\begin{figure}[!ht]
    \centering
    \begin{subfigure}{0.24\textwidth}
        \centering
        \includegraphics[width=\linewidth,trim=0pt 0pt 0pt 45pt,clip]{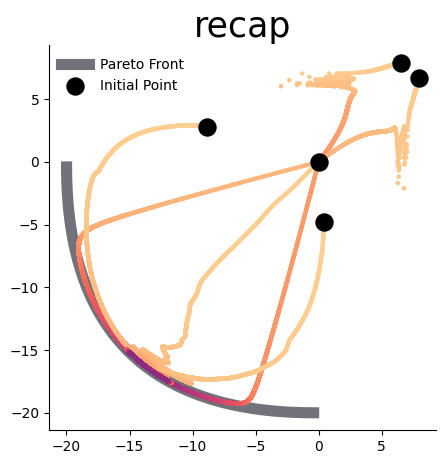}
        \caption{$T$ = 0.1 \label{fig:temp=.1}}
    \end{subfigure}
    \begin{subfigure}{0.24\textwidth}
        \centering
        \includegraphics[width=\linewidth,trim=0pt 0pt 0pt 45pt,clip]{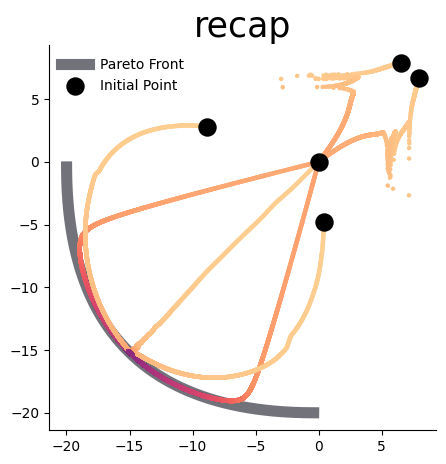}
        \caption{$T$ = 1.0}
    \end{subfigure}
        \begin{subfigure}{0.24\textwidth}
        \centering
        \includegraphics[width=\linewidth,trim=0pt 0pt 0pt 45pt,clip]{figure-new/mtl/recap.png}
        \caption{$T$  = 10.0}
    \end{subfigure}
    \begin{subfigure}{0.24\textwidth}
        \centering
        \includegraphics[width=\linewidth,trim=0pt 0pt 0pt 45pt,clip]{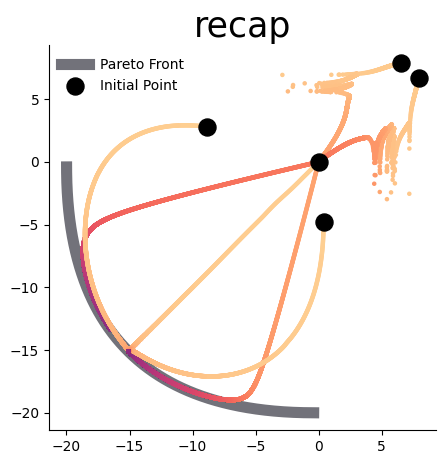}
        
        \caption{$T$ = 30.0 \label{fig:temp=30}}
    \end{subfigure}
    \caption{\textbf{Ablation studies on the temperature $T$.}  $T>1$ acts as a regularization to avoid extreme reweighting (one domain dominates others) and stabilizes the training.}
    \label{fig:temperature}
\end{figure}

For the temperature, setting $T$ too low makes training unstable: as shown in Figure~\ref{fig:temp=.1}, the trajectories exhibit strong fluctuations near the Pareto front. Conversely, setting $T$ to a high value (e.g., $T=30$ in Figure~\ref{fig:temp=30}) also harms convergence: for the two initializations farthest from the Pareto front, optimization requires many more steps to approach the front (the trajectories remain red for longer).

Due to the large computational requirements of RL training, we find that setting $T = 5$ and $\alpha = 0.5$ works reasonably well in the RLVR-only setting. For simplicity, we keep this configuration for the hybrid setup and do not perform additional hyperparameter tuning in the large-scale setting. Table~\ref{tab:small-ablation} reports the results when varying the trade-off $\alpha$, the temperature $T$, and the window size $W$. Although up-weighting the instability term can increase the weight assigned to the accuracy reward because this term fluctuates substantially, it comes at the cost of sacrificing essential perception skills. For example, $\alpha = 0.25$ improves performance on SAT and ScienceQA by 0.1\% and 0.6\%, respectively, but reduces ChartQA and InfoVQA performance by 3\%. Similar to our illustrative example, decreasing the temperature induces higher variation across tasks—for instance, it yields the highest score on MathVista while reducing SAT performance by 3.2\%.

\begin{table}[!ht]
\centering
\caption{\textbf{Benchmark performance in RLVR-only setting.}  Ablation results when varying the temperature and convergence rate-instability trade-off.
}
\label{tab:small-ablation}
\resizebox{\textwidth}{!}{
\begin{tabular}{l|cccccc}
\toprule
\textbf{Model} & \textbf{SAT} & \textbf{ScienceQA} & \textbf{MathVista (mini)} & \textbf{ChartQA} & \textbf{InfoVQA} & \textbf{MMMU} \\
\midrule
MoDoMoDo                          & 50.0  & 65.7  & 32.2  & {70.4} & 59.9 & 39.1  \\
\hline
 \rowcolor{carolinablue!60} {RECAP} & {55.2} & {71.6}  & {33.2}  & {70.4} & {60.8} & {42.4} \\
 {$\alpha$ = 0.50, T = 1.0, W=10} & 52.0 &  {71.6} &  33.4 & 68.1 & 58.5 & 40.4 \\
 {$\alpha$ = 0.75, T = 5.0, W=10} & 54.4 & 71.2  &  32.9 & 70.0 & 60.7 & 41.0 \\
 {$\alpha$ = 0.25, T = 5.0, W=10} & 55.3  &  72.2 & 33.7  &  66.1 & 56.8 & 39.3 \\
  {$\alpha$ = 0.50, T = 5.0, W=50} & 51.9  & 70.5  &  32.9 &  69.9 & 59.7  & 40.8 \\
\bottomrule
\end{tabular}
}
\end{table}